\documentclass[journal]{IEEEtran}
\usepackage{graphicx}
\usepackage{multirow}
\usepackage[table,xcdraw,dvipsnames]{xcolor}
\usepackage{longtable}
\usepackage{color}
\usepackage{xcolor,colortbl}
\usepackage{tikz}
\def\checkmark{\tikz\fill[scale=0.4](0,.35) -- (.25,0) -- (1,.7) -- (.25,.15) -- cycle;}
\usepackage{pifont}
\usepackage{comment}
\usepackage[activate]{microtype}
\usepackage[colorlinks=false]{hyperref}

\usepackage{authblk}
\usepackage{booktabs}

\usepackage{url}
\usepackage[dvipsnames]{xcolor}
\usepackage{soul}

\usepackage[separate-uncertainty = true,multi-part-units = repeat]{siunitx}
\usepackage{array,amsfonts,amsmath}
\usepackage{caption}
\usepackage{verbatim}
\ifCLASSOPTIONcompsoc
\usepackage{fancyhdr}
 \usepackage[nocompress]{cite}
\else
 \usepackage{cite}
\fi
\ifCLASSINFOpdf
\else
\fi
\begin{document}

\title{Transformers in Speech Processing: A Survey}

\author[1]{Siddique Latif\thanks{Email: siddique.latif@usq.edu.au}}
\author[2]{Aun Zaidi}
\author[3]{Heriberto Cuay\'ahuitl}
\author[4]{Fahad Shamshad}
\author[5]{Moazzam Shoukat}
\author[6]{Muhammad Usama}
\author[7]{Junaid Qadir}

\affil[1]{Queensland University of Technology (QUT), Australia}
\affil[2]{Information Technology University, Pakistan}
\affil[3]{University of Lincoln, UK}
\affil[4]{Mohamed bin Zayed University of Artificial Intelligence}
\affil[5]{EmulationAI}
\affil[6]{National University of Computer and Emerging Sciences, Pakistan}
\affil[7]{Qatar University, Doha}


\maketitle

\begin{abstract}

The remarkable success of transformers in the field of natural language processing has sparked the interest of the speech-processing community, leading to an exploration of their potential for modeling long-range dependencies within speech sequences. Recently, transformers have gained prominence across various speech-related domains, including automatic speech recognition, speech synthesis, speech translation, speech para-linguistics, speech enhancement, spoken dialogue systems, and numerous multimodal applications. In this paper, we present a comprehensive survey that aims to bridge research studies from diverse subfields within speech technology. By consolidating findings from across the speech technology landscape, we provide a valuable resource for researchers interested in harnessing the power of transformers to advance the field. We identify the challenges encountered by transformers in speech processing while also offering insights into potential solutions to address these issues.



\end{abstract}



\section{Introduction}
\label{sec:introduction}

Transformers have garnered significant attention in the speech processing and natural language processing (NLP) communities~\cite{karita2019comparative,lin2022survey,tay2022efficient,zhang2020transformer,wolf2019huggingface,wolf2020transformers} owing to their remarkable performance across a spectrum of applications, including  machine translation~\cite{wang2019learning},  automatic speech recognition (ASR)~\cite{dong2018speech,song2022multimodal}, question answering~\cite{yang2021just}, speech enhancement~\cite{dang2022dpt}, speech emotion recognition~\cite{wagner2022dawn}, and speech separation~\cite{subakan2021attention}, to name a few. These models have even surpassed traditional recurrent neural networks (RNNs), 
that struggle with long sequences and the vanishing gradient problem on sequence-to-sequence tasks~\cite{karita2019comparative}. The rapid development and popularity of transformer-based models in speech processing have generated a wealth of literature investigating the unique features that underlie their superior performance.

Transformers have an advantage in comprehending speech, as they analyze the entire sentence at once, whereas RNNs can only process smaller sections at a time. This is made possible by the unique self-attention-based architecture of transformers~\cite{vaswani2017attention}, which enables them to learn long-term dependencies, which is critical for speech processing tasks.
Moreover, the multi-head attention mechanism~\cite{voita2019analyzing}---a specialized feature in transformers---allows for more efficient parallelization during training, making them ideal for handling large datasets, which is a common challenge in speech processing tasks. This unique combination of self-attention and multi-head attention empowers transformers to achieve exceptional performance in sequence-to-sequence modeling, making them an indispensable tool for researchers and practitioners in the field of speech processing.

As the use of transformers for speech processing research community is gaining popularity, it is timely to review the existing literature and present a comprehensive overview of the field. In this regard, we provide a comprehensive overview of transformer model applications in the speech processing domain. Our aim is to assist researchers and practitioners in grasping the major trends and recent advancements in the field. Specifically, the main contributions of this survey are as follows:

\begin{itemize}
\item We present the first comprehensive survey of the application of transformer models in the speech processing field. Our survey covers more than 100 papers to cover the recent progress.

\item We provide detailed coverage of this rapidly evolving field by categorizing the papers based on their applications
Specifically, these applications include automatic speech recognition, neural speech synthesis, speech translation, speech enhancement, multi-modal applications, and spoken dialogue systems.

\item Finally, based on our thorough analysis, 
we identify various challenges and propose future research directions. Moreover, we provide valuable insights into potential solutions based on the literature reviewed.
\end{itemize}
We compare our paper with recent surveys on transformers and speech processing in Table \ref{table:lit-review}. It can be found that most of the transformer-related survey papers are focused on computer vision and natural language processing (NLP). The articles focused on speech processing do not cover transformers. Here, we focus on recent development in speech technology using transformers. Although other recent surveys have focused on deep learning 
techniques for  SR \cite{latif2023survey}, ASR \cite{zhang2018deep,nassif2019speech}, and SER \cite{khalil2019speech,latif2021survey}, none of
these has focused on transformers for speech processing. This
study bridges this gap by presenting an up-to-date survey of
research that focused on speech processing using transformers. The paper is organised as follows. Section \ref{sec:background} provides an overview of the applications of seq2seq models in SP and introduces the salient concepts underlying transformers. Section \ref{sec:applications} presents a comprehensive review of the applications of transformer models in SP. Section \ref{sec:challenges} discusses open problems and future research directions. Finally, in Section \ref{sec:conclusion}, we summarize and conclude the paper.

\begin{table*}[!ht]
\centering
\scriptsize
\caption{Comparison of this paper with other recent survey papers. Where Speech Processing = SP, Speech Emotion Recognition (SER), Natural Language Processing = NLP, Computer Vision = CV, Action Recognition = AR, and Reinforcement Learning = RL}
\begin{tabular}{|l|l|l|l|l|}
\hline
\textbf{Author (year)} & \textbf{\begin{tabular}[c]{@{}l@{}}Speech\\ Focused\end{tabular}} & \textbf{\begin{tabular}[c]{@{}l@{}}Transformer\\ Focused\end{tabular}} & \textbf{Domain}  & \textbf{Details}   \\ \hline
\begin{tabular}[c]{@{}l@{}}Deng et al.\\ 2016 \cite{deng2016deep}\end{tabular}    & \multicolumn{1}{c|}{\checkmark}   & \multicolumn{1}{c|}{$\times$}   & \begin{tabular}[c]{@{}l@{}}SP\end{tabular} & \begin{tabular}[c]{@{}l@{}}The paper reviews self-supervised representation learning methods in\\ speech processing, discussing generative, contrastive, and predictive\\ methods and multi-modal data approaches.\end{tabular} 
  \\ \hline
\begin{tabular}[c]{@{}l@{}}Khalil et al. \\2019 \cite{khalil2019speech}\end{tabular}    & \multicolumn{1}{c|}{\checkmark}   & \multicolumn{1}{c|}{$\times$}   & \begin{tabular}[c]{@{}l@{}}SER\end{tabular}   & \begin{tabular}[c]{@{}l@{}}This paper provides an overview of deep learning techniques for \\speech  based emotion recognition, covering databases used, emotions \\extracted, contributions made, and limitations related to it.\end{tabular}

    \\ \hline
\begin{tabular}[c]{@{}l@{}}Nassif et al. \\2019 \cite{nassif2019speech}\end{tabular}    & \multicolumn{1}{c|}{\checkmark}   & \multicolumn{1}{c|}{$\times$}   & \begin{tabular}[c]{@{}l@{}}SP\end{tabular} & \begin{tabular}[c]{@{}l@{}}This survey paper reviews the use of deep learning for speech-related\\ applications, providing a statistical analysis of 174 papers published\\ between 2006 and 2018.\end{tabular} \\ \hline
\begin{tabular}[c]{@{}l@{}}Alam et. al.\\ 2020 \cite{alam2020survey}\end{tabular} & \multicolumn{1}{c|}{$\times$}    & \multicolumn{1}{c|}{$\times$}   & Multimodal & \begin{tabular}[c]{@{}l@{}}The survey covers DNN architectures, algorithms, and systems for\\ speech and vision applications, not limited to transformers or speech.\end{tabular}  

   \\ \hline
\begin{tabular}[c]{@{}l@{}}Bracsoveanu et al.\\ 2020 \cite{bracsoveanu2020visualizing}\end{tabular} & \multicolumn{1}{c|}{$\times$}    & \multicolumn{1}{c|}{\checkmark}  & \begin{tabular}[c]{@{}l@{}}NLP\end{tabular}  & \begin{tabular}[c]{@{}l@{}}The survey paper focuses on explaining Transformer architectures\\ through visualizations to provide better understanding and proposes\\ a set of requirements for future Transformer visualization frameworks.\end{tabular}

   \\ \hline
\begin{tabular}[c]{@{}l@{}}Tan et al.\\ 2021 \cite{tan2021survey}\end{tabular} & \multicolumn{1}{c|}{\checkmark}   & \multicolumn{1}{c|}{$\times$}   & \begin{tabular}[c]{@{}l@{}}SP\end{tabular} & \begin{tabular}[c]{@{}l@{}}This paper provides a comprehensive survey on neural text-to-speech,\\ covering key components such as text analysis, acoustic models, \\ as well as advanced topics like fast, low-resource, robust, expressive, \\and adaptive TTS, and it also discusses future research directions.\end{tabular}  \\ \hline
\begin{tabular}[c]{@{}l@{}}Alharbi et al. \\2021 \cite{alharbi2021automatic}\end{tabular} & \multicolumn{1}{c|}{\checkmark}   & \multicolumn{1}{c|}{$\times$}   & \begin{tabular}[c]{@{}l@{}}ASR\end{tabular} & \begin{tabular}[c]{@{}l@{}}This survey paper provides a systematic review of automatic speech\\ recognition (ASR) technology, covering significant topics and recent\\ challenges published in the last six years.\end{tabular}  \\ \hline
\begin{tabular}[c]{@{}l@{}}Malik et al.\\ 2021 \cite{malik2021automatic}\end{tabular} & \multicolumn{1}{c|}{\checkmark}   & \multicolumn{1}{c|}{$\times$}   & \begin{tabular}[c]{@{}l@{}}ASR\end{tabular} & \begin{tabular}[c]{@{}l@{}}The survey paper compares various deep learning techniques and \\feature extraction methods for ASR and discusses the impact of \\different speech datasets on ASR performance, providing online \\resources and language models for ASR formulation.\end{tabular}

  \\ \hline
\begin{tabular}[c]{@{}l@{}}Liu et al.\\ 2021 \cite{liu2021survey}\end{tabular}    & \multicolumn{1}{c|}{$\times$}    & \multicolumn{1}{c|}{\checkmark}  & \begin{tabular}[c]{@{}l@{}}CV\end{tabular}   & \begin{tabular}[c]{@{}l@{}}The survey explores Transformer-based architectures in CV tasks, \\proposing a taxonomy and evaluating and comparing existing \\methods. It suggests three research directions for future investment.\end{tabular}

\\ \hline
\begin{tabular}[c]{@{}l@{}}Xu et al.\\ 2022 \cite{xu2022multimodal}\end{tabular}  & \multicolumn{1}{c|}{$\times$}    & \multicolumn{1}{c|}{\checkmark}  & Multimodal & \begin{tabular}[c]{@{}l@{}}The paper surveys Transformer techniques in multimodal learning, \\including theoretical reviews and applications. It aims to provide \\insights for researchers and practitioners.\end{tabular} \\ \hline
\begin{tabular}[c]{@{}l@{}}Lin et al. \\2022 \cite{lin2022survey}\end{tabular}    & \multicolumn{1}{c|}{$\times$}    & \multicolumn{1}{c|}{\checkmark}  & Multimodal & \begin{tabular}[c]{@{}l@{}}The survey reviews various Transformer variants in AI fields and \\proposes a new taxonomy. It covers architectural modifications, pre-\\training, applications, and potential directions for future research.\end{tabular} \\ \hline
\begin{tabular}[c]{@{}l@{}}Shamshad et al.\\ 2022 \cite{shamshad2022transformers}\end{tabular}  & \multicolumn{1}{c|}{$\times$}    & \multicolumn{1}{c|}{\checkmark}  & Medical Imaging  & \begin{tabular}[c]{@{}l@{}}The paper reviews Transformer models in medical imaging, discussing \\their applications and identifying open problems and future directions.\end{tabular} \\ \hline
\begin{tabular}[c]{@{}l@{}}Acheampong et al.\\ 2022 \cite{acheampong2021transformer}\end{tabular}   & \multicolumn{1}{c|}{$\times$}    & \multicolumn{1}{c|}{\checkmark}  & \begin{tabular}[c]{@{}l@{}}NLP\end{tabular}  & \begin{tabular}[c]{@{}l@{}}The paper reviews Transformer-based models used for NLP tasks in\\ emotion recognition, highlighting their strengths and limitations, and\\ providing future research directions.\end{tabular}    \\ \hline
\begin{tabular}[c]{@{}l@{}}Khan et al. \\2022 \cite{khan2022transformers}\end{tabular}  & \multicolumn{1}{c|}{$\times$}    & \multicolumn{1}{c|}{\checkmark}  & \begin{tabular}[c]{@{}l@{}}CV\end{tabular}   & \begin{tabular}[c]{@{}l@{}}The paper reviews Transformer models in CV tasks, covering a wide \\range of tasks and comparing the advantages and limitations of popular \\techniques. It also discusses research directions and future works.\end{tabular} \\ \hline
\begin{tabular}[c]{@{}l@{}}Tay et al.\\ 2022 \cite{tay2022efficient}\end{tabular} & \multicolumn{1}{c|}{$\times$}    & \multicolumn{1}{c|}{\checkmark}  & Multimodal & \begin{tabular}[c]{@{}l@{}}The article provides an overview of "X-former" models in multiple \\domains, aimed at improving efficiency and helping researchers \\navigate the evolving field.\end{tabular}    \\ \hline
\begin{tabular}[c]{@{}l@{}}Aleissaee et al. \\2022 \cite{aleissaee2022transformers}\end{tabular}    & \multicolumn{1}{c|}{$\times$}    & \multicolumn{1}{c|}{\checkmark}  & Remote Sensing   & \begin{tabular}[c]{@{}l@{}}The paper reviews transformers-based methods for remote sensing\\ problems. It also discusses different challenges and open issues.\end{tabular} \\ \hline
\begin{tabular}[c]{@{}l@{}}Ulhaq et al. \\2022 \cite{ulhaq2022vision}\end{tabular}  & \multicolumn{1}{c|}{$\times$}    & \multicolumn{1}{c|}{\checkmark}  & AR & \begin{tabular}[c]{@{}l@{}}It reviews literature on vision transformer techniques for action \\recognition, providing taxonomies, network learning strategies, \\and evaluation metrics, while discussing challenges and future \\research directions.\end{tabular}  \\ \hline
\begin{tabular}[c]{@{}l@{}}Bhangale et al. \\2022 \cite{bhangale2022survey}\end{tabular}  & \multicolumn{1}{c|}{\checkmark}   & \multicolumn{1}{c|}{$\times$}   & \begin{tabular}[c]{@{}l@{}}SP\end{tabular} & \begin{tabular}[c]{@{}l@{}}The paper presents a survey of deep learning techniques for various\\ speech processing applications. It covers various deep learning models\\ such as Auto-Encoder, GAN, RBN, DBN, DNN, CNN, RNN, and \\DRL, along with speech databases and evaluation metrics.\end{tabular}  
  \\ \hline
\begin{tabular}[c]{@{}l@{}}Lahoud et al. \\2023 \cite{lahoud20223d}\end{tabular}  & \multicolumn{1}{c|}{$\times$}    & \multicolumn{1}{c|}{\checkmark}  & 3D Vision  & \begin{tabular}[c]{@{}l@{}}It reviews over 100 transformer methods on various 3D CV tasks,\\ comparing their performance to common non-transformer methods\\ on 3D benchmarks. It also discusses open directions and challenges.\end{tabular}

\\ \hline
\begin{tabular}[c]{@{}l@{}}Li et al. \\2023 \cite{li2023survey}\end{tabular}  & \multicolumn{1}{c|}{$\times$}    & \multicolumn{1}{c|}{\checkmark}  & \begin{tabular}[c]{@{}l@{}}RL\end{tabular}  & \begin{tabular}[c]{@{}l@{}}The paper reviews recent advances and applications of transformers in\\ reinforcement learning, providing a taxonomy of existing works in the\\ field and summarizing future prospects.\end{tabular}  \\ \hline
This paper & \multicolumn{1}{c|}{\checkmark}   & \multicolumn{1}{c|}{\checkmark}  & \begin{tabular}[c]{@{}l@{}} SP\end{tabular} & \begin{tabular}[c]{@{}l@{}}
The paper reviews applications of transformers and challenges faced \\in various speech processing applications, like speech recognition, \\synthesis, translation, and enhancement, and suggests future research\\ directions for improving speech technology with transformers.

\end{tabular} \\ \hline
\end{tabular}
\label{table:lit-review}
\end{table*}

\section{Background} \label{sec:background}

In this section, we will provide a comprehensive overview of transformer architecture, starting with a brief overview of sequential models and their limitations in handling sequential data. We will then delve into the key concepts behind the transformer's operation, highlighting the unique features that enable it to outperform traditional recurrent neural networks. Lastly, we will discuss popular transformers for speech processing.


\subsection{Sequential Models for Speech Processing} 


Early deep learning approaches in the SP domain typically employed variants of convolutional neural networks (CNNs)~\cite{abdel2014convolutional,zhang2017towards}. However, a drawback of these CNN-based approaches is their inability to capture the sequential nature of speech data. This limitation of CNNs led to the development of sequence-to-sequence (seq2seq) architectures, such as RNNs and long short-term memory networks (LSTMs), which are specifically designed for sequential data. RNNs are well-suited for sequential data because they can process long sequences step-by-step with limited memory of previous sequence elements. More recently, researchers have also combined the strengths of CNNs and RNNs by using CNNs to extract audio features and using these features as input to train RNNs. However, RNNs have been shown to have issues with the vanishing or exploding gradient problem. To address this issue, LSTMs use a gating mechanism and memory cells to control the flow of information and alleviate gradient problems. Many LSTM variations---such as Frequency-LSTM, Time-Frequency LSTMs, Bi-directional LSTMs, ConvLSTMs, and Stacked LSTMs---have been proposed for SP tasks. Despite their effectiveness, seq2seq models are limited in important ways: they cannot take advantage of parallel computing hardware and have difficulties modeling long-term context.


\subsection{Overview of Transformers} 

Transformers were first introduced in the seminal work of Vaswani et al. titled ``Attention Is All You Need'' \cite{vaswani2017attention} for machine translation tasks in Natural Language Processing (NLP) and have recently shown impressive performance in other domains, including computer vision, medical imaging, and remote sensing. Transformer models use self-attention layers to effectively capture long-range dependencies among the input sequence, which is in contrast to traditional recurrent neural networks that struggle to capture such interaction. Furthermore, self-attention allows for more parallelization compared to recurrent neural networks, as these can process the speech sequence as a whole without relying on past states to capture dependencies. More specifically, two types of attention were introduced by Vaswani et al. including (1) scaled dot-product attention, and (2) multi-head attention. In addition, positional encoding is also used to inject information about the relative or absolute position of the tokens in the sequence. Due to these desirable properties, transformers have garnered immense interest in the speech community, and several approaches have been proposed that build upon transformers. We provide next a brief overview of the core components of transformers, which are multi-head self-attention layers and a position-wise feed-forward network (positional encoding). 


\vspace{2mm}
\subsubsection{Self-Attention Layer}

Self-attention (SA) layer aims to capture the internal correlation of sequence or features by aggregating global information from the entire input sequence, a task that is challenging for conventional recurrent models. 
Specifically, the input $\boldsymbol{X} \in \mathbb{R}^{N \times D}$ consisting of $N$ entities each having dimension $D$ is transformed into query ($\boldsymbol{Q} \in \mathbb{R}^{N \times D_k}$), key ($\boldsymbol{K} \in \mathbb{R}^{N \times D_k}$) and value ($\boldsymbol{V} \in \mathbb{R}^{N \times D_k}$) matrices via learnable weight matrices $\boldsymbol{W}^{Q} \in \mathbb{R}^{D \times D_k}$, $\boldsymbol{W}^{K} \in \mathbb{R}^{D \times D_k}$, and $\boldsymbol{W}^{V} \in \mathbb{R}^{D \times D_k}$ respectively. Mathematically,
\begin{align}
 \boldsymbol{Q} = \boldsymbol{X}\boldsymbol{W}^{Q}, \quad
 \boldsymbol{K} = \boldsymbol{X}\boldsymbol{W}^{K}, \quad
 \boldsymbol{V} = \boldsymbol{X}\boldsymbol{W}^{V}.
\end{align}

\begin{figure*}[!t]
\centering
\includegraphics[width=0.9\textwidth]{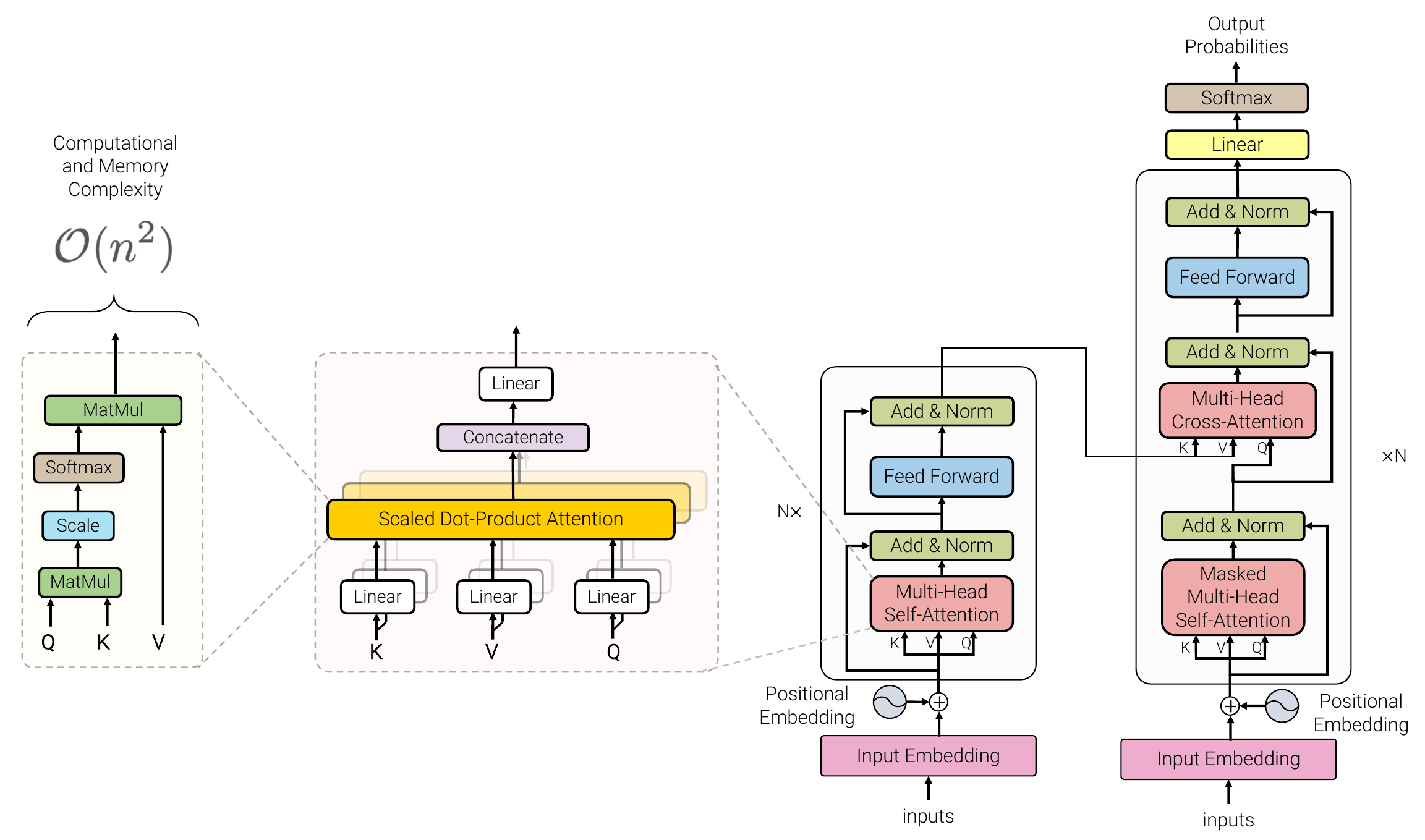}
\caption{Architecture of the standard transformer (Adapted from Vaswani et al., 2017 \cite{vaswani2017attention} and Tay et al., 2020 \cite{tay2022efficient}). \textit{The model comprises encoder and decoder layers, each with stacked self-attention and feed-forward sub-layers. The encoder produces hidden states from an input token sequence, while the decoder generates predictions from an output token sequence and attends to the encoder's states for input information.}}
\label{fig:lc}
\end{figure*}

Then the dot-product of the query matrix $\boldsymbol{Q}$ with all the keys $\boldsymbol{Q}$ in a given sequence is computed and the resulting matrix is normalized using the softmax operator to get the attention matrix $\boldsymbol{A} \in \mathbb{R}^{N \times N}$ as
\begin{equation}
 \boldsymbol{A} = \textit{softmax} \left( \frac{\boldsymbol{QK}^{T}}{\sqrt{D_k}}\right)\boldsymbol{V}.
\end{equation}

The output of the SA layer $\boldsymbol{Z}$ is the attention matrix $\boldsymbol{A}$ multiplied by the value matrix $\boldsymbol{V}$
\begin{equation}
 \boldsymbol{Z} = \boldsymbol{AV}.
\end{equation}

\vspace{2mm}
\subsubsection{Masked Self-Attention} In the original transformers paper \cite{vaswani2017attention}, the SA blocks used in the decoder are masked to prevent attending to the subsequent future entities by element-wise multiplication of the mask $\boldsymbol{M} \in \mathbb{R}^{N \times N}$ as: 
\begin{equation}
 \boldsymbol{Z} = \textit{softmax} \left( \frac{\boldsymbol{QK}^{T}}{\sqrt{d_k}} \circ \boldsymbol{M}\right)\boldsymbol{V},
\end{equation}
where $\boldsymbol{M}$ is the upper triangular matrix and $\circ$ denotes the Hadamard product. This is called masked self-attention.

\vspace{2mm}
\subsubsection{Multi-Head Attention}
Rather than only computing the attention once, Multi-Head Self-Attention (MHSA) consists of multiple SA blocks. These SA blocks are concatenated together channel-wise to model dependencies among different elements in the input sequence. Each head in MHSA has its own learnable weight matrices denoted by $\{\mathbf{W}^{Q_i},\mathbf{W}^{K_i},\mathbf{W}^{V_i} \}$, where $i=0 \cdots (h{-}1)$ and $h$ denotes total number of heads in MHSA block. Specifically,
$$\text{MHSA}(\mathbf{Q},\mathbf{K},\mathbf{V}) = [\mathbf{Z}_0,\mathbf{Z}_1,...,\mathbf{Z}_{h-1}]\mathbf{W}^{O},$$
whereas $\mathbf{W}^{O} \in \mathbb{R}^{h.D_k \times N}$ computes linear transformation of heads and $\mathbf{Z}_i$ can be written as,
$$ \mathbf{Z}_i = \textit{softmax}\left (\frac{\mathbf{QW}^{Q_i}(\mathbf{KW}^{K_i})^{T}}{\sqrt{D_k/h}}\right ) \mathbf{VW}^{V_i}.$$

\vspace{2mm}
\subsubsection{Positional Encoding}

The self-attention mechanism in transformer models allows for input speech frames to be processed in no particular order or position. To account for this, positional encoding is used to provide the transformer model with information about the order of the input sequence. This is done by associating each position in the input sequence with a vector that helps the transformer learn positional relationships. Positional encoding can be learned during training or pre-defined and can be encoded in relative or absolute ways for SP tasks.

\subsection{Popular Transformers for Speech}

Transformers are a novel neural network architecture that relies solely on attention mechanisms to handle sequential data, such as natural language and speech, for various tasks. Since the seminal work of Vaswani et al. (2017) \cite{vaswani2017attention}, many extensions and applications of transformers have been developed for natural language processing (NLP) tasks, such as language modeling, question answering, sentiment analysis, text generation, etc. Transformers are also becoming increasingly popular in the speech community due to their suitability for various tasks including speech recognition, enhancement, text-to-speech synthesis, speaker recognition, and multi-microphone processing. Various open-source libraries including Hugging Face, SpeechBrain, and torch audio are accelerating the research in the speech domain. Big tech companies like Google, Meta, Amazon, etc., are building large speech domain-related transformer models. 
While transformers were initially developed for NLP tasks, they have since been adapted for other data types, including speech. BERT \cite{devlin2018bert} is a language model that uses masked-language modeling (MLM) as its pre-training objective. BERT consists of two modules: an embedding layer that maps tokens to vectors, and an encoder layer that applies self-attention and feed-forward networks to learn contextualized representations. While BERT and similar text-based large language models (e.g., GPT \cite{radford2018improving}), XLNet \cite{yang2019xlnet}, T5 \cite{raffel2020exploring}), etc.) have been successful in various NLP tasks, their application to speech processing is limited due to several shortcomings. For instance, they require discrete tokens as input, which means it needs a tokenizer or a speech recognition system to convert raw audio signals into text, introducing errors and noise in the process that can negatively impact the performance \cite{wu2020sentence}. Additionally, these models are pre-trained on large-scale text corpora, which may not match the domain or style of speech data, leading to domain mismatch issues. 


\subsubsection{wav2vec}
To overcome these limitations, dedicated frameworks for learning speech representations, such as wav2vec, have been developed. wav2vec uses a self-supervised training approach that leverages the contrastive predictive coding (CPC) loss function to learn speech representations without the need for transcription or segmentation \cite{baevski2020vq,baevski2020wav2vec}. This approach allows wav2vec to achieve state-of-the-art performance on several speech processing tasks, including speech recognition, speaker recognition, and spoken language understanding, among others \cite{baevski2020vq,baevski2020wav2vec}. 
w2v-BERT is a framework that combines contrastive learning and MLM for self-supervised speech pre-training and builds on the success of wav2vec. w2v-BERT consists of three modules: a feature encoder, a quantization module, and a masked prediction module. The feature encoder is similar to wav2vec, but the quantization module discretizes the continuous speech representations into a finite set of speech units using Gumbel-softmax. 
The main differences between wav2vec and w2v-BERT lie in their pre-training objectives and the data types they operate on. wav2vec focuses on raw audio signals and uses contrastive learning to learn speech representations. In contrast, w2v-BERT operates on discrete tokens and it uses both MLM and contrastive learning as pre-training objectives. 
 After the success of the wav2vec models, Baevski et al. demonstrated that it could be fine-tuned with a small amount of labeled data to achieve state-of-the-art results on speech recognition tasks \cite{pepino2021emotion,novoselov2022robust}. This breakthrough led to the development of a series of models aimed at building cross-lingual or multilingual speech recognition systems using pre-trained transformer models. 
Another model is XLS-R \cite{babu2021xls}, a large-scale model for cross-lingual speech representation learning based on wav2vec 2.0. XLS-R is trained on nearly half a million hours of publicly available speech audio in 128 languages and achieves state-of-the-art results on a wide range of tasks, domains, data regimes, and languages. These models leverage large-scale multilingual data augmentation and contrastive learning techniques to learn universal speech representations that can be transferred across languages and domains.

\subsubsection{data2vec}

While the success of wav2vec was inspired by the achievements of BERT, it paved the way for the development of other dedicated frameworks that use transformers to learn representations from multi-modal data. One such example is data2vec, which aims to learn multi-modal representations of data, including speech, images, and text, using a contrastive learning objective \cite{baevski2022data2vec}. Similar to wav2vec, data2vec uses a self-supervised training approach that does not require labels or annotations and learns representations by maximizing agreement between differently augmented views of the same data sample. However, unlike wav2vec, which focuses solely on speech signals, data2vec can operate on various types of data and can learn joint representations that capture cross-modal correlations and transfer knowledge across modalities. 
Data2vec's self-supervised training approach allows it to learn representations without the need for labeled data, making it a scalable and cost-effective solution for many applications. Data2vec has been shown to outperform other unsupervised approaches for learning multimodal representations, such as Skip-thought \cite{kiros2015skip} and Paragraph Vector \cite{le2014distributed}, on several benchmark datasets \cite{baevski2022data2vec}. However, it should be noted that while data2vec is suitable for learning representations that generalize across domains and modalities, it may not perform as well as domain-specific models for certain tasks, such as speech recognition or speaker identification, where the data has a specific domain or language.

\subsubsection{Whisper}

Whisper \cite{radford2022robust} is a general-purpose model designed for speech recognition in noisy or low-resource settings, and is capable of performing multiple speech-related tasks.  Whisper uses weak supervision and a minimalist approach to data pre-processing. It achieves state-of-the-art results, showcasing the potential of using advanced machine-learning techniques in speech processing.
Whisper is capable of performing multilingual speech recognition, speech translation, and language identification. It is trained on a large dataset of diverse audio and is a multitasking model that can handle various speech-related tasks, such as transcription, voice assistants, education, entertainment, and accessibility. 
The Whisper model is unique in that it uses a minimalist approach to data pre-processing, allowing models to predict the raw text of transcripts without significant standardization. This eliminates the need for a separate inverse text normalization step to produce naturalistic transcriptions, simplifying the speech recognition pipeline. The resulting models can generalize well to standard benchmarks and are competitive with prior fully supervised results without fine-tuning. 


\subsubsection{Tacotron}

Transformer-based models have also gained popularity in speech synthesis tasks. A prime example of such a model is Tacotron \cite{wang2017tacotron}, which uses a sequence-to-sequence architecture with attention mechanisms to generate high-quality speech from text input. The limitations of the Griffin-Lim algorithm used for audio signal generation led to the development of Tacotron 2 \cite{shen2018natural} by Google AI in 2018, which used WaveNet to generate raw audio waveforms directly from mel-spectrograms, resulting in more natural-sounding speech. Furthermore, Microsoft introduced Transformer TTS \cite{li2019neural} in 2019, which employs a transformer network instead of the convolutional and recurrent networks used in Tacotron 2, along with a duration predictor and a new training method that uses teacher forcing for faster convergence and better performance. Despite these advancements, current systems still have limitations in generating natural-sounding speech for non-English languages, handling complex intonations and accents, and real-time speech synthesis for applications such as voice assistants and automated phone systems.

\begin{table*}[!ht]
\begin{center}
\scriptsize
\caption{Transformer Speech Models: Release Year and Parameter Count with Task Compatibility. * means that the parameters for this model cannot be found or are proprietary.}
\begin{tabular}{|l|l|l|lll|l|}
\hline
\multirow{2}{*}{\textbf{Model Name}} & \multicolumn{1}{c|}{\multirow{2}{*}{\textbf{\begin{tabular}[c]{@{}c@{}}Release\\ Year\end{tabular}}}} & \multicolumn{1}{c|}{\multirow{2}{*}{\textbf{\begin{tabular}[c]{@{}c@{}}Number of\\ Parameters\end{tabular}}}} & \multicolumn{3}{c|}{\textbf{Tasks}}                                                                                                                                                                                                                                                                           & \multicolumn{1}{c|}{\multirow{2}{*}{\textbf{Multimodal}}} \\ \cline{4-6}
                                     & \multicolumn{1}{c|}{}                                                                                 & \multicolumn{1}{c|}{}                                                                                         & \multicolumn{1}{c|}{\textbf{\begin{tabular}[c]{@{}c@{}}Speech\\ Synthesis (TTS)\end{tabular}}} & \multicolumn{1}{c|}{\textbf{\begin{tabular}[c]{@{}c@{}}Speech\\ Translation (ST)\end{tabular}}} & \multicolumn{1}{c|}{\textbf{\begin{tabular}[c]{@{}c@{}}Automatic Speech\\ Recognition (ASR)\end{tabular}}} & \multicolumn{1}{c|}{}                                     \\ \hline
Tacotron \cite{wang2017tacotron}    & 2017  &  13 million \cite{wang2019deep} & \multicolumn{1}{c|}{\checkmark}   & \multicolumn{1}{c|}{$\times$}  & \multicolumn{1}{c|}{$\times$}  & \multicolumn{1}{c|}{$\times$}   \\ \hline
Tacotron 2 \cite{shen2018natural}   & 2017  & 28.2 million \cite{beliaev2021talknet}  & \multicolumn{1}{c|}{\checkmark}   & \multicolumn{1}{c|}{$\times$}  & \multicolumn{1}{c|}{$\times$}  & \multicolumn{1}{c|}{$\times$}    \\ \hline
Transformer-TTS \cite{li2019neural}    & 2018  & 30.7 million \cite{ren2019fastspeech}  & \multicolumn{1}{c|}{\checkmark}   & \multicolumn{1}{c|}{$\times$}  & \multicolumn{1}{c|}{$\times$}  & \multicolumn{1}{c|}{$\times$} \\ \hline
vq-wav2vec \cite{baevski2020vq} & 2019  &  34 million\cite{bgn2021timeline} & \multicolumn{1}{c|}{$\times$}    & \multicolumn{1}{c|}{$\times$}  & \multicolumn{1}{c|}{\checkmark} & \multicolumn{1}{c|}{$\times$} \\ \hline
Mockingjay \cite{liu2019mockingjay} & 2019  &  85 million\cite{bgn2021timeline} & \multicolumn{1}{c|}{$\times$}    & \multicolumn{1}{c|}{$\times$}  & \multicolumn{1}{c|}{\checkmark} & \multicolumn{1}{c|}{$\times$} \\ \hline
FastSpeech  \cite{ren2019fastspeech} & 2019  &  23 million \cite{ren2020fastspeech} & \multicolumn{1}{c|}{\checkmark}   & \multicolumn{1}{c|}{$\times$}  & \multicolumn{1}{c|}{$\times$}  & \multicolumn{1}{c|}{$\times$}    \\ \hline
wav2vec \cite{schneider2019wav2vec}  & 2019  & 16 million \cite{chen2022speechformer}  & \multicolumn{1}{c|}{$\times$}    & \multicolumn{1}{c|}{$\times$}  & \multicolumn{1}{c|}{\checkmark}  & \multicolumn{1}{c|}{$\times$}    \\ \hline
wav2vec 2.0 \cite{baevski2020wav2vec} & 2020  & 317 million\cite{bgn2021timeline}  & \multicolumn{1}{c|}{$\times$}    & \multicolumn{1}{c|}{$\times$}  & \multicolumn{1}{c|}{\checkmark}  & \multicolumn{1}{c|}{$\times$}    \\ \hline
FastSpeech 2 \cite{ren2020fastspeech} & 2020  & 27 million \cite{ren2020fastspeech}  & \multicolumn{1}{c|}{\checkmark}   & \multicolumn{1}{c|}{$\times$}  & \multicolumn{1}{c|}{$\times$}  & \multicolumn{1}{c|}{$\times$}    \\ \hline
FastPitch \cite{lachowicz2020fastpitch}   & 2020  &  26.8 million & \multicolumn{1}{c|}{\checkmark}   & \multicolumn{1}{c|}{$\times$}  & \multicolumn{1}{c|}{$\times$}  & \multicolumn{1}{c|}{$\times$} \\ \hline
Conformer \cite{gulati2020conformer}  & 2020  & 1 billion\cite{bgn2021timeline}  & \multicolumn{1}{c|}{$\times$}    & \multicolumn{1}{c|}{$\checkmark$} & \multicolumn{1}{c|}{\checkmark}  & \multicolumn{1}{c|}{$\times$}    \\ \hline
DeCoAR 2.0 \cite{chang2020decoar}   & 2020  & 317 million\cite{bgn2021timeline}  & \multicolumn{1}{c|}{$\times$}    & \multicolumn{1}{c|}{$\times$}  & \multicolumn{1}{c|}{\checkmark}  & \multicolumn{1}{c|}{$\times$} \\ \hline
w2v-Conformer  & 2021  &  1 billion\cite{bgn2021timeline} & \multicolumn{1}{c|}{$\times$}    & \multicolumn{1}{c|}{$\times$}  & \multicolumn{1}{c|}{\checkmark}  & \multicolumn{1}{c|}{$\times$}    \\ \hline
w2v-BERT \cite{wang2019bridging}    & 2021  & 1 billion\cite{bgn2021timeline}  & \multicolumn{1}{c|}{$\times$}    & \multicolumn{1}{c|}{$\times$}  & \multicolumn{1}{c|}{\checkmark}  & \multicolumn{1}{c|}{$\times$} \\ \hline
HuBERT \cite{hsu2021hubert}& 2021  & 317 million\cite{bgn2021timeline}  & \multicolumn{1}{c|}{$\times$}    & \multicolumn{1}{c|}{$\times$}  & \multicolumn{1}{c|}{\checkmark}  & \multicolumn{1}{c|}{$\times$}   \\ \hline
XLS-R \cite{babu2021xls}& 2021  & 2 billion\cite{bgn2021timeline}  & \multicolumn{1}{c|}{$\times$}    & \multicolumn{1}{c|}{\checkmark} & \multicolumn{1}{c|}{\checkmark}  & \multicolumn{1}{c|}{$\times$}    \\ \hline
UniSpeech \cite{chen2021unspeech}   & 2021  & 317 million\cite{bgn2021timeline}  & \multicolumn{1}{c|}{$\times$}    & \multicolumn{1}{c|}{$\times$}  & \multicolumn{1}{c|}{\checkmark}  & \multicolumn{1}{c|}{$\times$} \\ \hline
UniSpeech-SAT  \cite{liu2021unispeechsat} & 2021  &  317 million\cite{bgn2021timeline}  & \multicolumn{1}{c|}{$\times$}    & \multicolumn{1}{c|}{$\times$}  & \multicolumn{1}{c|}{\checkmark}  & \multicolumn{1}{c|}{$\times$} \\ \hline
BigSSL \cite{hu2021bigssl} & 2021  & 8 billion\cite{bgn2021timeline}  & \multicolumn{1}{c|}{$\times$}    & \multicolumn{1}{c|}{$\times$}  & \multicolumn{1}{c|}{\checkmark} & \multicolumn{1}{c|}{$\times$} \\ \hline
WavLM \cite{liu2021wavlm}  & 2021  & 317 million\cite{bgn2021timeline}  & \multicolumn{1}{c|}{$\times$}    & \multicolumn{1}{c|}{$\times$}  & \multicolumn{1}{c|}{\checkmark}  &\multicolumn{1}{c|}{$\times$} \\ \hline
DeltaLM \cite{liu2021deltalm} & 2021  &  360 million \cite{ma2021deltalm} & \multicolumn{1}{c|}{\checkmark}   & \multicolumn{1}{c|}{\checkmark} & \multicolumn{1}{c|}{$\times$} & \multicolumn{1}{c|}{$\checkmark$} \\ \hline
SpeechT5 \cite{ao2021speecht5}& 2021  & \multicolumn{1}{c|}{11 billion \cite{ao2021speecht5}}  & \multicolumn{1}{c|}{\checkmark}   & \multicolumn{1}{c|}{\checkmark} & \multicolumn{1}{c|}{\checkmark} & \multicolumn{1}{c|}{$\checkmark$}   \\ \hline
data2vec \cite{baevski2022data2vec} & 2022  & \multicolumn{1}{c|}{*}  & \multicolumn{1}{c|}{$\times$}    & \multicolumn{1}{c|}{$\times$}  & \multicolumn{1}{c|}{\checkmark} & \multicolumn{1}{c|}{$\checkmark$}    \\ \hline
data2vec 2.0 \cite{baevski2022efficient}  & 2022  &  \multicolumn{1}{c|}{*} & \multicolumn{1}{c|}{$\times$}    & \multicolumn{1}{c|}{$\times$}  & \multicolumn{1}{c|}{\checkmark} & \multicolumn{1}{c|}{$\checkmark$}    \\ \hline
SpeechFormer \cite{chen2022speechformer}    & 2022  &  \multicolumn{1}{c|}{3.5 million \cite{chen2022speechformer}} & \multicolumn{1}{c|}{$\times$}    & \multicolumn{1}{c|}{$\times$} & \multicolumn{1}{c|}{$\checkmark$}  & \multicolumn{1}{c|}{\checkmark}   \\ \hline
Whisper \cite{radford2022robust}    & 2022  &  \multicolumn{1}{c|}{1.6 billion \cite{openai2022whisper}} & \multicolumn{1}{c|}{$\times$}    & \multicolumn{1}{c|}{\checkmark} & \multicolumn{1}{c|}{\checkmark}  & \multicolumn{1}{c|}{$\times$}   \\ \hline
VALL-E \cite{chen2022valle}& 2023  & \multicolumn{1}{c|}{*}  & \multicolumn{1}{c|}{\checkmark}   & \multicolumn{1}{c|}{$\times$}  & \multicolumn{1}{c|}{$\times$}  & \multicolumn{1}{c|}{$\times$}    \\ \hline
VALL-E X \cite{zhang2023speak}& 2023  &  \multicolumn{1}{c|}{*} & \multicolumn{1}{c|}{\checkmark}   & \multicolumn{1}{c|}{$\times$}  & \multicolumn{1}{c|}{$\times$}  & \multicolumn{1}{c|}{$\times$}    \\ \hline
\end{tabular}
\label{table:transformer-models}
\end{center}
\end{table*}

\subsubsection{VALL-E}
VALL-E \cite{chen2022valle} is another model that has gained attention, a zero-shot text-to-speech synthesis system that uses a language modeling approach, treating TTS as a conditional language modeling task rather than continuous signal regression. It is trained using discrete codes from an off-the-shelf neural audio codec model and pre-trained on 60,000 hours of English speech data, providing strong in-context learning capabilities. Unlike previous TTS systems, VALL-E does not require additional structure engineering, pre-designed acoustic features, or fine-tuning. It can synthesize high-quality personalized speech with only a 3-second acoustic prompt from an unseen speaker. The model also provides diverse outputs with the same input text and can preserve the acoustic environment and the speaker's emotion of the acoustic prompt. VALL-E's speaker dimension is built on a generalized TTS system, leveraging a large amount of semi-supervised data. This approach is significant, as scaling up semi-supervised data has been underestimated for TTS. Evaluation results show that VALL-E significantly outperforms the state-of-the-art zero-shot TTS system on LibriSpeech and VCTK datasets in terms of speech naturalness and speaker similarity. VALL-E X \cite{zhang2023speak} was developed as a natural extension of VALL-E to address the challenge of cross-lingual speech synthesis. Cross-lingual speech synthesis involves generating speech in a target language, using a source language speech prompt and a target language text prompt. While VALL-E was designed for zero-shot TTS in English, there was a need for a model that could handle cross-lingual speech synthesis in multiple languages. VALL-E X, therefore, extends VALL-E to support cross-lingual synthesis by training a multi-lingual model to predict acoustic token sequences in the target language using prompts in both source and target languages. This enables the model to generate high-quality speech in the target language while preserving the voice, emotion, and acoustic environment of the unseen speaker and effectively alleviates the foreign accent problem, which can be controlled by a language ID.



\subsubsection{Conformer}
Recent advances in transformers for speech processing have also seen the emergence of the Conformer models \cite{gulati2020conformer}. The Conformer architecture combines convolutional and transformer layers, enabling it to capture both local and global context information. This makes Conformer models well-suited for speech-processing tasks such as speech recognition and speaker identification, where capturing long-range dependencies is crucial. Conformer achieved state-of-the-art performance on benchmarks such as LibriSpeech and AISHELL-1. However, previous limitations in speech synthesis and recognition, such as the struggle to produce natural-sounding speech in languages other than English and generate speech in real-time, remained. In response, Wang et al. \cite{zhang2020pushing} presented an ASR model that uses a combination of noisy student training with SpecAugment and giant Conformer models pre-trained using the wav2vec 2.0 pre-training method on the Libri-Light dataset. They achieved state-of-the-art word error rates on the LibriSpeech dataset. In 2021, Wang et al. \cite{wang2022conformer} extended Conformer and developed Conformer-LHUC, which utilized learning hidden unit contribution (LHUC) for speaker adaptation. Conformer-LHUC showed superior performance in elderly speech recognition and has potential implications for clinical diagnosis and treatment of Alzheimer's disease. 

\subsubsection{UniSpeech}
There are other emerging models that are gaining traction in speech processing, such as the UniSpeech model, which focuses on developing models that can handle low-resource and cross-lingual speech tasks. Microsoft's UniSpeech approach \cite{wang2021unispeech} proposes a unified pre-training method that combines supervised and unsupervised learning for speech representation learning. The approach uses phonetic CTC learning and phonetically-aware contrastive self-supervised learning to capture more phonetic information and generalize better across languages and domains. The authors evaluate UniSpeech on cross-lingual representation learning and achieve state-of-the-art results on low-resource speech recognition tasks. 
In a related paper \cite{chen2022unispeech}, the authors propose UniSpeech-SAT, a universal speech representation learning method with speaker-aware pre-training. The method improves existing self-supervised learning for speaker representation learning by using utterance-wise contrastive learning and utterance mixing augmentation. The method achieves state-of-the-art performance in universal representation learning, especially for speaker identification tasks, and can be easily adapted to downstream tasks with minimal fine-tuning. 

\subsubsection{Speechformer}
After the success of UniSpeech models, the field of end-to-end speech recognition has continued to advance. In June 2021, Speechformer \cite{papi2021speechformer}, a self-supervised pre-trained model for end-to-end speech recognition that leverages masked acoustic modeling and contrastive predictive coding was proposed. Unlike previous models that used either convolutional or recurrent neural networks, Speechformer uses a transformer-based encoder-decoder architecture with relative position encoding and layer normalization. It is pre-trained on 53k hours of unlabeled speech data and achieves competitive results on several ASR benchmarks. 

\subsubsection{WavLM}
Microsoft Research Asia released WavLM \cite{chen2022wavlm}, a large-scale self-supervised pre-trained model that can solve full-stack downstream speech tasks such as ASR, TTS, and speaker verification. WavLM jointly learns masked speech prediction and denoising in pre-training and employs gated relative position bias for the Transformer structure to better capture the sequence ordering of the input speech. It is trained on a massive dataset of 94k hours of speech and achieves state-of-the-art results on several downstream speech tasks for 10 languages. 

There are various other transformers for speech-related tasks that we present in Table \ref{table:transformer-models}.

\section{Literature Review} \label{sec:applications}

\subsection{Automatic Speech Recognition (ASR)}

ASR enables machines to recognize uttered speech and transform it into the corresponding sequence of text (words or sub-words). State-of-the-art ASR systems achieved improved performance by using RNNs with long short-term memory (LSTM) \cite{schmidhuber1997long} units as their backbone networks. Recently, there has been an increasing interest in the exploitation of transformers \cite{vaswani2017attention} for ASR, inspired by their success in different NLP tasks such as language modeling \cite{dai2019transformer} and machine translation \cite{vaswani2018tensor2tensor}. RNNs process the input signal in a sequential manner by utilizing expensive back-propagation through time (BPTT) \cite{werbos1990backpropagation} algorithm to learn temporal dependencies. Transformers circumvent this with a self-attention mechanism to capture the temporal correlations among the sequential data. This enables transformers to capture longer temporal correlations with less computation complexity. Another advantage of using a transformer is the ability to parallelize the computations in transformers, which can reduce the time for training deeper models on larger datasets. 

In ASR, transformers achieved a competitive recognition rate compared to RNN-based baseline models. For instance, Karita et al. \cite{karita2019comparative} experimentally compared transformers with conventional RNNs. Based on the results, they showed various training and performance benefits achieved with transformers in comparison to RNNs. In \cite{zeyer2019comparison}, Zeyer et al. performed a comparison of the transformer encoder decoder-attention model with RNNs and found that transformers are more stable compared to LSTM, however, they face the problem of overfitting and generalization. They also found that the pretraining leads to faster convergence and performance boost. Li et al. \cite{li2020comparison} performed a comparison between RNN and transformer-based end-to-end models using the 65 thousand hours of Microsoft
anonymized training data. They found that transformer-based attention encoder-decoder architecture achieved the best accuracy. Similarly, studies \cite{wang2020transformer,zhou2018comparison} also performed a comparison of transformers with different ASR systems and highlights the benefits of transformers and pointers for future research.

In hybrid ASR, an acoustic encoder is used to encode an input sequence to high-level embedding vectors that are exploited to generate the posterior distribution of tied states of the hidden Markov model (HMM). Combined with other knowledge sources, these posterior distributions are used to construct a search graph. A decoder network is then used to determine the best hypothesis. Different deep models can be used as acoustic encoders in hybrid ASR. Recently, studies started using transformers for improving hybrid acoustic modeling. Wang et al. \cite{wang2020transformer} evaluated a transformer-based acoustic model for hybrid speech recognition. They explored multiple modeling choices and losses for training deep transformers. Based on the results, they showed that the proposed hybrid ASR can achieve significantly improved WER compared to the very strong bi-directional LSTM (BLSTM) baselines. 

For streaming applications of ASR, Wu et al. \cite{wu2020streaming} presented an acoustic model based on an augmented memory self-attention transformer for hybrid ASR. The proposed model attends a short segment of the input sequence and accumulates information into memory banks. This makes the segment information equally accessible. Evaluations were performed on Librispeech data, which showed that the proposed model achieves a 15\% error reduction in contrast to the widely used LC-BLSTM baseline.

Recurrent sequence-to-sequence models have achieved great progress in ASR. These models are based on the encoder-decoder architecture, where the encoder transforms the speech feature sequence into hidden representations and generates an output sequence. Conventional RNNs-based sequence-to-sequence models suffer from slow training and training parallelization issues. In a transformer-based sequence-to-sequence model, the encoder and decoder network are composed of multi-head attention and position-wise feed-forward networks rather than RNNs. Also, the encoder outputs are attended by each decoder block respectively. This makes training transformer-based sequence-to-sequence models faster and allows for parallel training. Dong et al. \cite{dong2018speech} presented a Speech-Transformer with no recurrence to learn positional dependencies in speech signals entirely relying on attention mechanisms. Evaluations were performed on Wall Street Journal (WSJ) dataset, which showed that transformers can achieve a competitive word error rate (WER), significantly faster than the published results using RNN-based sequence-to-sequence models. Zhou et al. \cite{zhou2018comparison} explored the modeling units in ASR using transformer-based sequence-to-sequence models on Mandarin Chinese speech. They performed a comparison among five modeling units including context-independent phonemes, syllables, words, sub-words, and characters. Based on the results, they found that the character-based model performs best and achieves state-of-the-art (SoTA) CER on the HKUST dataset. 


In a study by Zhou et al. \cite{zhou2018syllable}, the authors compared the performance of a context-independent (CI)-phoneme-based model and a syllable-based model using a transformer on the HKUST dataset. The results showed that the syllable-based model performed better than the CI-phoneme-based model. In Hrinchuk et al. \cite{hrinchuk2020correction}, a transformer-based sequence-to-sequence model was proposed to improve the performance of automatic speech recognition (ASR) by correcting the ASR system output. The proposed model was able to correct erroneous outputs into semantically and grammatically correct text, which helped improve the performance. To address the issue of asynchronous encoding and decoding in sequence-to-sequence models, Tain et al. \cite{tian2020synchronous} presented a synchronous transformer that can predict the output in chunks. The experiments showed that the proposed model was able to encode and decode synchronously, which led to an improved character error rate (CER) rate.

Transformers have also shown promising results in large-scale ASR. Lu et al. \cite{lu2020exploring} explored transformers for large-scale ASR with 65,000 hours of training data. They investigated different aspects such as warm-up training, model initialisation, and layer normalization techniques on scaling up transformers for ASR. Chen et al. \cite{chen2020developing} evaluated the potential of transformer Transducer models for the first pass decoding with low latency and fast speed on a large-scale ASR dataset. Based on the experiments, they showed that the Transformer Transducer model outperforms RNN Transducer (RNN-T) \cite{graves2012sequence}, streamable transformer, and hybrid model in the streaming scenario. In \cite{li2019speechtransformer}, Li et al. focused on a large-scale Mandarin ASR and propose three optimization strategies to improve the efficiency and performance of SpeechTransformer \cite{dong2018speech}. Wang et al. \cite{wang2020transformer} performed a comparative study on the transformer-based acoustic model on large-scale ASR. They found that the transformer-based ASR model achieves better performance compared to LSTM for voice assistant
tasks. The aforementioned studies show the effectiveness of transformers for ASR. We summarise recent studies on transformers for ASR in Table \ref{tab:ASR-Studies}. 

\subsection{Neural Speech Synthesis}

Neural speech synthesis, or Neural text-to-speech (TTS), is an important field of research that aims to synthesize speech from text input. Traditional TTS systems are composed of complex components including acoustic frontends, duration model, acoustic prediction model, and vocoder models \cite{taylor2009text}. The complexity of the TTS has been recently overcome with deep end-to-end TTS architectures \cite{wang2017tacotron,arik2017deep}. These systems can synthesize realistic-sounding speech by training on $<$text,audio$>$ pairs, and eliminate the need for complex sub-components and their separate training. Prominent models includes Tacotron \cite{wang2017tacotron}, Tacotron 2 \cite{shen2018natural}, Deep Voice 3 \cite{ping2018deep}, and Clarinet \cite{ping2018clarinet}. These models generate Mel-spectrogram from text input, which is then used to synthesize speech by vocoder such as Griffin-Lim \cite{griffin1984signal}, WaveNet \cite{vanwavenet}, and Waveglow \cite{prenger2019waveglow}. 

Recently, transformers are becoming popular to generate Mel-spectrogram in TTS systems. Particularly, they replace RNN structures in end-to-end TTS to improve training and inference efficiency. In \cite{li2019neural}, Li et al. attempted to utilize the multi-head attention mechanism to replace RNN structures as well as the vanilla attention mechanism in Tacotron 2. This helps in improving pluralization by solving the long-distance dependency problem. They generated the Mel-spectrogram using the phoneme sequences as input and exploited WaveNet as a vocoder to synthesize speech samples. Based on the results, they showed that transformer TTS was able to speed up training 4.25 times compared to Tacotron 2 and achieve a similar MOS performance.

\begin{table*}[!ht]
\centering
\caption{ Recent studies on transformers for \textbf{Automatic Speech Recognition (ASR)}.}
\scriptsize
\begin{tabular}{|l|l|l|l|}
\hline
Author (year) & Dataset   & Performance   & Architecture  \\ \hline
Zeyer et al. \cite{zeyer2019comparison} &\begin{tabular}[c]{@{}l@{}}LibriSpeech (1000 hr), \\TED-LIUM 2 (200 hr) \\ and Switchboard (300 hr)\end{tabular} & \begin{tabular}[c]{@{}l@{}}WER\\LibriSpeech: 2.81\% \\TED LIUM: 12.0\% \\Switchboard: 10.6 \% \end{tabular} & \begin{tabular}[c]{@{}l@{}}Transformer encoder-decoder-attention model and\\ LSTM encoder-decoder-attention model.\end{tabular} 
 \\ \hline
Li et al. \cite{li2020comparison} & \begin{tabular}[c]{@{}l@{}} Microsoft transcribed\\data (65000 hr)\end{tabular} & WER: 9.16\% & \begin{tabular}[c]{@{}l@{}} Recurrent neural network transducer (RNN-T),\\ RNN attention-based encoder-decoder (AED), and \\Transformer-AED \end{tabular} \\ \hline
 
 Wang et al. \cite{wang2020transformer} & \begin{tabular}[c]{@{}l@{}} Personal Assistant Dataset\\ \end{tabular} &\begin{tabular}[c]{@{}l@{}}
 WER: 3.94 \end{tabular} & \begin{tabular}[c]{@{}l@{}} Transformer, \\Emformer (streamable variant of Transformer), \\latency-controlled BLSTM (LCBLSTM), and\\ LSTM \end{tabular} \\ \hline

Wu et al \cite{wu2020streaming} &
\begin{tabular}[c]{@{}l@{}}LibriSpeech and \\ German and Russian \\ video dataset \end{tabular} 
 & \begin{tabular}[c]{@{}l@{}}WER\\LibriSpeech: 2.8\% 
 \\Russian: 18.0\% \\German: 17.4\% \end{tabular} & \begin{tabular}[c]{@{}l@{}}Streaming Transformer with self-attention with \\augmented memory (SAAM) module.\end{tabular} \\\hline

Dong et al. \cite{dong2018speech} & \begin{tabular}[c]{@{}l@{}}WSJ dataset\end{tabular} & WER: 10.9\% & \begin{tabular}[c]{@{}l@{}}Transformer encoder-decoder-attention model with\\ 2D-Attention mechanism.\end{tabular} \\ \hline

Zhou et al. \cite{zhou2018syllable} & HKUST Dataset & CER: 28.77\% & \begin{tabular}[c]{@{}l@{}}Transformer encoder-decoder-attention model with \\syllable-based input and output.\end{tabular} 
\\ \hline
Hrinchuk et al. \cite{hrinchuk2020correction} &
\begin{tabular}[c]{@{}l@{}} Prepared own dataset \\ 101K sample for first name and\\ 580K sample for last name \end{tabular} & \begin{tabular}[c]{@{}l@{}}WER: 9.2\% on first \\ and 6.6\% on last name\end{tabular} & \begin{tabular}[c]{@{}l@{}}Transformer encoder-decoder-attention model for \\ASR post-processing.\end{tabular} \\\hline
Tian et al. \cite{tian2020synchronous} & AIShell& \begin{tabular}[c]{@{}l@{}}CER: 8.91\% \end{tabular} & \begin{tabular}[c]{@{}l@{}} Synchronous Transformer (a variant of Transformer \\with chunk-by-chunk prediction)
\end{tabular} \\ \hline

Lu et al. \cite{lu2020exploring} & Microsoft data (65000 hr) & \begin{tabular}[c]{@{}l@{}}WER: 12.2\% \end{tabular} & \begin{tabular}[c]{@{}l@{}}Transformer encoder-decoder\end{tabular} \\ \hline

Chen et al. \cite{chen2020developing} & Microsoft data (65000 hr) & \begin{tabular}[c]{@{}l@{}}WER
: 8.19\% \\\end{tabular} & \begin{tabular}[c]{@{}l@{}}Transformer transducer\end{tabular} \\ \hline

Li et al. \cite{li2019speechtransformer} & \begin{tabular}[c]{@{}l@{}}AiShell 1(165 hr) \\ HKUST(156 hr)\end{tabular} & \begin{tabular}[c]{@{}l@{}}CER\\AiShell-1: 13.09\% \\ HKUST: 28.95\%\end{tabular} &
\begin{tabular}[c]{@{}l@{}} 
SpeechTransformer
\end{tabular} \\ \hline

Lancucki et al. \cite{lancucki2020fastpitch} & \begin{tabular}[c]{@{}l@{}} LJSpeech-1.1 Dataset \end{tabular} & \begin{tabular}[c]{@{}l@{}}
MOS Values\\
4.071±0.164
\end{tabular} & \begin{tabular}[c]{@{}l@{}}	FastPitch, a variant of FastSpeech \end{tabular} \\ \hline

Mohammed et al. \cite{mohamed2019transformers} &
\begin{tabular}[c]{@{}l@{}} LJSpeech-1.1 Dataset \end{tabular}& \begin{tabular}[c]{@{}l@{}}WER: 4.7\% \end{tabular}& \begin{tabular}[c]{@{}l@{}}Transformer encoder-decoder with convolutional\\ context modules \end{tabular} \\ \hline

Zhang et al. \cite{zhang2021transmask} &
\begin{tabular}[c]{@{}l@{}} SWBD\\AMI\\AISHELL\\ \end{tabular}& \begin{tabular}[c]{@{}l@{}}WER: \\SWBD: 7.1\%\\AMI: 24.1\%\\AISHELL: 4.7\% \end{tabular}& \begin{tabular}[c]{@{}l@{}}TransMask, a transformer encoder-decoder with\\ mask prediction heads\end{tabular} \\ \hline

Moriya et al.\cite{moriya2020self} &
\begin{tabular}[c]{@{}l@{}} WER: WSJ, Switchboard, \\Librispeech, CSJ and \\NTT Japanese dataset. \end{tabular}& \begin{tabular}[c]{@{}l@{}}WER: \\SWBD: 8.9\%\\LibriSpeech: 4.4\%\\CER:\\CSJ: 3.9\% \\NTT: 4.2\% \end{tabular}& \begin{tabular}[c]{@{}l@{}}CTC-Transformer, a transformer encoder-decoder \\with connectionist temporal classification (CTC) loss\end{tabular} \\ \hline

Cao et al.\cite{cao2021improving} &
\begin{tabular}[c]{@{}l@{}} LibriSpeech \end{tabular}& \begin{tabular}[c]{@{}l@{}}WER: 3.5\%\end{tabular}& \begin{tabular}[c]{@{}l@{}}Streaming Transformer, a transformer encoder-decoder\\ with block processing and latency control mechanisms\end{tabular} \\ \hline

Tsunoo et al.\cite{tsunoo2019transformer} &
\begin{tabular}[c]{@{}l@{}}
WSJ, Librispeech, \\VoxForge Italian, and\\ AISHELL-1
 \end{tabular}& \begin{tabular}[c]{@{}l@{}}
WER:\\
Librispeech: 4.6\%\\
WSJ: 5.7\%\\
AISHELL-1: 7.6\%\\
VoxForge: 10.3\%\end{tabular}& \begin{tabular}[c]{@{}l@{}}Transformer encoder-decoder with contextual block\\ processing (CBP), a technique to improve streaming ASR \\performance by using past and future context information\end{tabular} \\ \hline

Jain et al.\cite{jain2020finnish} &
\begin{tabular}[c]{@{}l@{}}
YLE news dataset
 \end{tabular}& \begin{tabular}[c]{@{}l@{}}
WER:\\
17.71 \% \end{tabular}& \begin{tabular}[c]{@{}l@{}}Transformer encoder-decoder with deep self-\\attention layers\end{tabular} \\ \hline

Yu et al.\cite{yu2020dual} &
\begin{tabular}[c]{@{}l@{}}
LibriSpeech and MultiDomain

 \end{tabular}& \begin{tabular}[c]{@{}l@{}}
WER:\\LibriSpeech: 2.5\%\\
MultiDomain: 6.0\% 

\end{tabular}& \begin{tabular}[c]{@{}l@{}}Dual-mode ASR (DM-ASR), a hybrid model that \\combines streaming ASR (S-ASR) and full-context ASR \\(F-ASR) using two parallel transformer encoders and one\\ shared decoder.\end{tabular} \\ \hline

\end{tabular}
\label{tab:ASR-Studies}
\end{table*}

In order to improve the inference speed, FastSpeech \cite{ren2019fastspeech} used a feed-forward network based on 1D convolution \cite{gehring2017convolutional,jin2018fftnet} and the self-attention mechanism in transformers to generate Mel-spectrogram in parallel. It utilizes the length regulator based on duration predictor to solve the issue of sequence length mismatch between the Mel-spectrogram sequence and its corresponding phoneme sequence. Fastspeech was evaluated on the LJSpeech dataset and results showed that it can significantly speed up the generation of Mel-spectrogram while achieving comparable performance to the autoregressive transformer model. FastPitch \cite{lancucki2020fastpitch} improves FastSpeech by conditioning the TTS model on fundamental frequency or pitch contour. Pitch conditioning improved the convergence and removed the requirement for knowledge distillation of Mel-spectrogram targets in FastSpeech. Fastspeech 2 \cite{ren2020fastspeech} is another transformer-based TTS system that resolved the issues in Fastspeech and better addressed the one-to-many mapping problem in TTS. It uses more diverse information of speech (e.g., energy, pitch, and more accurate duration) as conditional inputs and directly train the system on a ground-truth target. Fastspeech 2s is another variant proposed in \cite{ren2020fastspeech}, which further simplifies the speech synthesis pipeline by directly generating speech from the text in inference without using Mel-spectrograms as intermediate output. Experiments on the LJSpeech data showed that FastSpeech 2 and FastSpeech 2s present a simplified training pipeline with fast, robust, and controllable speech synthesis compared to FastSpeech.

End-to-end TTS systems, such as FastSpeech \cite{ren2019fastspeech} and Durian \cite{yu2019durian}, utilize a duration model to align output acoustic features with the input text. However, the multi-stage training pipeline used in these systems can be slow. To address this issue, Lim et al. proposed a jointly trained duration-informed transformer (JDI-T) that uses a feed-forward transformer with a duration predictor to generate acoustic feature sequences without explicit alignments. JDI-T achieved state-of-the-art performance on the Korean Single Speaker Speech (KSS) dataset and synthesized high-quality speech compared to other popular TTS models. 

However, neural TTS models can suffer from robustness issues and generate poor audio samples for unseen or unusual text. To overcome these issues, Li et al. proposed RobuTrans, a robust transformer that converts input texts to linguistic features before feeding them to the encoder \cite{li2020robutrans}. They also modified the attention mechanism and position embedding to improve the learning of holistic information from the input, resulting in improved MOS scores compared to other popular TTS models. Another approach to achieving robustness in TTS systems is the segment-transformer (s-Transformer) \cite{wang2020s} proposed by Wang et al. The s-Transformer is capable of modeling speech at the segment level, allowing it to capture long-term dependencies and use segment-level encoder-decoder attention to handle long sequence pairs. This approach enables the s-Transformer to achieve similar performance to the standard transformer while also exhibiting robustness on extra-long sentences. Lastly, Zheng et al. \cite{zheng2020improving} proposed an approach that incorporates a local recurrent neural network into the transformer to capture both sequential and local information in sequences. Evaluation on a 20-hour Mandarin speech corpus demonstrated that this model outperforms the transformer alone in terms of performance.

Speech synthesis using multi-speaker voices is another interesting field of research. Chen et al. presented a MultiSpeech model, based on transformer TTS, that can synthesize high-quality speech in multi-speaker voices with fast inference speed. To achieve this, they designed a special component in the transformer to preserve positional information and prevent copy between consecutive speech frames. The MultiSpeech model was evaluated on VCTK and LibriTTS datasets, and the results demonstrated superior performance compared to existing models. Voice conversion, on the other hand, focuses on altering the source speaker's voice to match the target voice without changing the linguistic content. While various studies have explored RNN-based sequence-to-sequence models for voice conversion, these models require extensive training data and often suffer from mispronunciation issues. To address these challenges, Huang et al. presented a voice transformer network that utilized pre-training to improve data-efficient training and achieve better results compared to RNN-based models. Recent studies have continued to push the boundaries of speech synthesis systems, exploring various approaches to improve performance. The summary of recent studies on speech synthesis is presented in Table \ref{tab:SpeechSynthesis-Studies}.

\begin{table*}[!ht]
\scriptsize
\caption{Recent studies on transformers for \textbf{Speech Synthesis}.}
\centering
\begin{tabular}{|l|l|l|l|}
\hline
\multicolumn{1}{|c|}{Author (year)} & \multicolumn{1}{c|}{Datasets} & Performance& \multicolumn{1}{c|}{Architecture}\\ \hline
\hline
Ren et al. \cite{ren2019fastspeech} & \begin{tabular}[c]{@{}l@{}} LJSpeech dataset \end{tabular} & \begin{tabular}[c]{@{}l@{}}
MOS: 3.84 ± 0.08
\end{tabular} & \begin{tabular}[c]{@{}l@{}}FastSpeech, a feed-forward Transformer TTS \end{tabular}\\ \hline

Ren et al. \cite{ren2020fastspeech} & \begin{tabular}[c]{@{}l@{}} LJSpeech dataset \end{tabular} & \begin{tabular}[c]{@{}l@{}}
MAE: \\FastSpeech2:0.131 \\FastSpeech2s:0.133
\end{tabular} &\begin{tabular}[c]{@{}l@{}} FastSpeech 2, an improved FastSpeech with\\ more variance information\end{tabular} \\ \hline

Chen et al. \cite{chen2020multispeech} & \begin{tabular}[c]{@{}l@{}} VCTK and LibriTTS \\datasets \end{tabular} & \begin{tabular}[c]{@{}l@{}}
MOS: \\VCTK: 3.65 ± 0.14 \\ LibriTTS: 2.95 ± 0.14 
\end{tabular} &\begin{tabular}[c]{@{}l@{}} MultiSpeech, a multi-speaker Transformer TTS\\ with speaker embeddings and classifier loss\end{tabular} \\ \hline

{\L}a{\'n}cucki et al. \cite{lancucki2020fastpitch} & \begin{tabular}[c]{@{}l@{}} LJSpeech Dataset \end{tabular} & \begin{tabular}[c]{@{}l@{}}
 MOS: 3.707 ± 0.218 
\end{tabular} &\begin{tabular}[c]{@{}l@{}} FastPitch, a FastSpeech variant with pitch\\ prediction and control\end{tabular} \\ \hline

Gehringi et al. \cite{gehring2017convolutional} & \begin{tabular}[c]{@{}l@{}} WMT'14 English-French \\
WMT'14 English-German.\\WMT'14 English-Romanian. \end{tabular} & \begin{tabular}[c]{@{}l@{}}
WMT'14 English-French: 20.51\\
WMT'14 English-German: 26.43\\WMT'14 English-Romanian: 41.62
\end{tabular} &\begin{tabular}[c]{@{}l@{}} ConvS2S, a CNN-based \\sequence-to-sequence model. \end{tabular} 
\\ \hline
Yu et al. \cite{yu2019durian} & \begin{tabular}[c]{@{}l@{}} Hours of Speech: \\
Male speaker: 18 hours\\
Female speaker: 7 hours\\
 \end{tabular} & \begin{tabular}[c]{@{}l@{}}
Male: 4.11 \\Female: 4.26
\end{tabular} &\begin{tabular}[c]{@{}l@{}} DurIAN, a multimodal TTS model with \\duration-informed attention network (DIAN) \end{tabular} \\ \hline

Lim et al. \cite{lim2020jdi} & \begin{tabular}[c]{@{}l@{}} Internal Speaker Dataset\\ Korean Speaker Dataset \end{tabular} & \begin{tabular}[c]{@{}l@{}}
MOS Values on a scale of 5\\
Internal: 3.77\\
KSS: 3.52
\end{tabular} & \begin{tabular}[c]{@{}l@{}}JDI-T, a feed-forward Transformer TTS with\\ duration predictor \end{tabular} \\ \hline

Wang et al. \cite{wang2020s} & \begin{tabular}[c]{@{}l@{}} Professional enUS\\ A speaker dataset (46 hour) \end{tabular} & \begin{tabular}[c]{@{}l@{}}
MOS Values on a scale of 5\\
Short: 4.29\\
Long: 4.2\\
Extra Long: 3.99
\end{tabular} & \begin{tabular}[c]{@{}l@{}}s-Transformer, a segment-wise Transformer \\TTS\end{tabular} \\ \hline

Zheng et al. \cite{zheng2020improving} & \begin{tabular}[c]{@{}l@{}} Mandarin speech corpus \end{tabular} & \begin{tabular}[c]{@{}l@{}}
MOS Values \\
4.34±0.05
\end{tabular} & \begin{tabular}[c]{@{}l@{}}LRN-Transformer, a Transformer TTS with \\LRNs\end{tabular} \\ \hline

Huang et al. \cite{huang2020voice} & \begin{tabular}[c]{@{}l@{}} CMU Arctic dataset \end{tabular} & \begin{tabular}[c]{@{}l@{}}
WER: 7.8\% \\
CER: 4.8\%

\end{tabular} & \begin{tabular}[c]{@{}l@{}}VTN, a seq2seq voice conversion model with\\ TTS pretraining \end{tabular} \\ \hline

Hu et al. \cite{hu2020unsupervised} & 
\begin{tabular}[c]{@{}l@{}} LibriTTS\\ VCTK \end{tabular} & \begin{tabular}[c]{@{}l@{}}
WER:
LibriTTS: 33.3 ± 1.2
\\ VCTK: 20.3 ± 1.2
\end{tabular} & \begin{tabular}[c]{@{}l@{}}MI-TTS, an unsupervised TTS model with \\VAEs and mutual information minimization \end{tabular} \\ \hline
Chen et al. \cite{chen2022fine} & 
\begin{tabular}[c]{@{}l@{}} LJSpeech\\ VCTK \end{tabular} & \begin{tabular}[c]{@{}l@{}}
WER:
LJSpeech: 9.5\%
\\ VCTK: 12.5 \%
\end{tabular} & \begin{tabular}[c]{@{}l@{}}TransformerTTS with local style tokens (LST) \\and cross-attention blocks \end{tabular} \\ \hline

Liu et al. \cite{liu2021graphspeech} & 
\begin{tabular}[c]{@{}l@{}} LJSpeech database\end{tabular} & \begin{tabular}[c]{@{}l@{}}
RMSE: 1.625\%

\end{tabular} & \begin{tabular}[c]{@{}l@{}}GraphSpeech, a graph neural network (GNN)\\ based TTS that encodes syntactic information \\as a dependency graph \end{tabular} \\ \hline

Wang et al. \cite{wang2021patnet} & 
\begin{tabular}[c]{@{}l@{}} LJSpeech database\end{tabular} & \begin{tabular}[c]{@{}l@{}}
CER: 1.7\%

\end{tabular} & \begin{tabular}[c]{@{}l@{}}PatNet, a phoneme-level autoregressive \\Transformer (TTS) that predicts\\ mel-spectrograms from phoneme sequences\end{tabular} \\ \hline

\end{tabular}
\label{tab:SpeechSynthesis-Studies}
\end{table*}

\begin{table*}[!ht]
\scriptsize
\caption{Recent studies on transformers for \textbf{Speech Translation}.}
\centering
\begin{tabular}{|l|l|l|l|}
\hline
\textbf{Author (year)} & \textbf{Datasets} & \textbf{Performance} & \textbf{Architecture(s)} \\ \hline
\begin{tabular}[c]{@{}l@{}}Kano et al.\\ 2021 \cite{kano2021transformer}\end{tabular} & \begin{tabular}[c]{@{}l@{}}Fisher Spanish-English,\\ LibriSpeech English-French,\\ LibriSpeech English-German\end{tabular}  & \begin{tabular}[c]{@{}l@{}}BLEU:\\ 16.9 (es-en), 15.8 (en-fr), 10.1 (en-de);\\ MOS:\\ 3.87 ± 0.08 (es-en), 3.81 ± 0.09 (en-fr), \\3.77 ± 0.09 (en-de);\\ MAE:\\ 0.131 (es-en), 0.132 (en-fr), 0.133 (en-de)\end{tabular} & \begin{tabular}[c]{@{}l@{}}Encoder-Decoder with attention and transcoder \\ module\end{tabular}  \\ \hline
\begin{tabular}[c]{@{}l@{}}Zhang et al.\\ 2019 \cite{zhang2019lattice}\end{tabular} & \begin{tabular}[c]{@{}l@{}}IWSLT 2017 en-de ST task,\\ MuST-C en-de ST task\end{tabular}  & \begin{tabular}[c]{@{}l@{}}BLEU: 17.9 on IWSLT 2017 en-de;\\ BLEU: 20.8 on MuST-C en-de\end{tabular} & \begin{tabular}[c]{@{}l@{}}A novel controllable lattice attention mechanism \\ that leverages the extra information from the\\ lattice structure of ASR output\end{tabular}  \\ \hline
\begin{tabular}[c]{@{}l@{}}Huawei et al.\\ 2022 \cite{huawei2022transformer}\end{tabular} & \begin{tabular}[c]{@{}l@{}}WMT 2014 English-German5,\\ WMT 2014 English-French5,\\ WMT 2016 Romanian-English5,\\ WMT 2016 English-Czech5,\\ WMT 2017 English-Turkish\end{tabular} & \begin{tabular}[c]{@{}l@{}}BLEU score:\\ 28.4 (en-de),\\ 41.0 (en-fr),\\ 32.8 (ro-en),\\ 26.7 (en-cs),\\ 24.9 (en-tr)\end{tabular} & Encoder-decoder with self-attention layers \\ \hline
\begin{tabular}[c]{@{}l@{}}Ao et al.\\ 2021 \cite{ao2021speecht5}\end{tabular}  & \begin{tabular}[c]{@{}l@{}}LibriSpeech,\\ LibriTTS,\\ Common Voice,\\ LJSpeech,\\ AISHELL-1/2/3,\\ VCTK-Corpus,\\ VoxCeleb1/2\end{tabular}    & \begin{tabular}[c]{@{}l@{}}WER: \\ 2.9\% (LibriSpeech test-clean), \\7.0\% (LibriSpeech test-other),\\ 6.8\% (AISHELL-1), \\6.0\% (AISHELL-2), \\7.9\% (AISHELL-3);\\ MOS:\\ 4.06 (LibriTTS), 4.01 (LJSpeech);\\ BLEU:\\ 17.8 (English-to-Chinese ST);\\ MCD:\\ 6.38 dB (voice conversion);\\ PESQ: 2.97 (speech enhancement);\\ EER: 0.83\% (speaker identification)\end{tabular} & \begin{tabular}[c]{@{}l@{}}Shared encoder-decoder network with six\\ modal-specific pre/post-nets, pre-trained with\\ contrastive loss, masked speech/text modeling \\loss and cross-modal alignment loss\end{tabular} \\ \hline
\end{tabular}
\label{tab:SpeechTranslation-Studies}
\end{table*}

\subsection{Speech Translation (ST)}

Speech Translation (ST) is the process of translating the spoken speech in the source language into the target language. ST systems are typically divided into two categories: cascaded systems and end-to-end systems. Cascaded ST systems consist of an automatic speech recognition (ASR) system and a machine translation (MT) system. ASR system generates text from the spoken sentence, which is then used by a machine translation system to translate it into the target language. Cascaded ST systems face the problem of errors compounding between components, e.g., recognition errors leading to larger translation errors. In contrast, end-to-end ST systems optimize a single model that directly translates the spoken utterance into the target language. Various studies explored methods and techniques to improve the performance of both cascaded ST systems \cite{ney1999speech,matusov2005integration,do2017toward} as well as end-to-end ST systems \cite{berard2016listen,berard2018end,weiss2017sequence}. 

Presently, ST research explores transformers for solving different issues. Vila et al. \cite{vila2018end} used a transformer for end-to-end speech translation. Evaluations were performed on the Spanish-to-English translation task and a bilingual evaluation understudy score was computed, which showed that the end-to-end architecture was able to outperform the concatenated systems. Zhang et al. \cite{zhang2019lattice} presented a lattice transformer for speech translation, which also uses lattice representation in addition to the traditional sequential input. They evaluated the proposed model on Spanish-English Speech Translation Corpus and achieved improvements over strong baseline results. In \cite{di2019adapting}, the authors present an adaptation of the transformer to end-to-end ST. They performed down-sampling of input with convolutional neural networks to i) make the training process feasible on GPUs, ii) model the bi-dimensional nature of a spectrogram, and iii) add a distance penalty to the attention, so as to bias it towards the local context. Furthermore, several distinct studies (Jia et al., 2021 \cite{jia2021translatotron}; Zhang et al., 2023 \cite{zhang2023speak}; Huang et al., 2022 \cite{huang2022transpeech}; Li et al., 2020 \cite{li2020multilingual}; Wang et al., 2020 \cite{wang2020fairseq}; Zeng et al., 2021 \cite{zeng2021realtrans}) explore various Speech Translation model implementations or models designed for Speech Translation tasks.


\begin{table*}
\scriptsize
\caption{Recent studies on transformers for \textbf{Speech Paralinguistics}.}
\centering
\begin{tabular}{|l|l|l|l|}
\hline
\textbf{Author (year)}  & \textbf{Datasets} & \textbf{Performance}  & \textbf{Architecture(s)}    \\ \hline
\begin{tabular}[c]{@{}l@{}}Chen et. al.\\ 2022 \cite{chen2022wavlm}\end{tabular}   & \begin{tabular}[c]{@{}l@{}}SUPERB benchmark,\\ LibriSpeech (ASR),\\ VoxCeleb1/2 (SV),\\ AMI (SD),\\ CommonVoice (LID),\\ CREMA-D (SER)\end{tabular} & \begin{tabular}[c]{@{}l@{}}SUPERB: 0.9231234,\\ LibriSpeech WER: 1.9/4.81,\\ VoxCeleb1/2 EER: 0.69/0.831,\\ AMI DER: 7.51,\\ CommonVoice LID ACC: 0.9781,\\ CREMA-D SER UAR: 0.7221\end{tabular}    & \begin{tabular}[c]{@{}l@{}}Transformer with gated relative position bias\\ and utterance mixing.\end{tabular}   \\ \hline
\begin{tabular}[c]{@{}l@{}}Shor et al.\\ 2022 \cite{shor2022trillsson}\end{tabular}  & \begin{tabular}[c]{@{}l@{}}SUPERB benchmark,\\ LibriSpeech (ASR),\\ VoxCeleb1/2 (SV),\\ AMI (SD),\\ CommonVoice (LID),\\ CREMA-D (SER)\end{tabular} & \begin{tabular}[c]{@{}l@{}}SUPERB: 0.906123,\\ LibriSpeech WER: 2.4/5.81,\\ VoxCeleb1/2 EER: 0.75/0.911,\\ AMI DER: 8.71,\\ CommonVoice LID ACC: 0.9761,\\ CREMA-D SER UAR: 0.713\end{tabular}  & \begin{tabular}[c]{@{}l@{}}EfficientNet-B0 with Audio Spectrogram Transformer \\encoder and ResNet-50 encoder for fixed-length and \\arbitrary-length inputs respectively.\end{tabular} \\ \hline
\begin{tabular}[c]{@{}l@{}}Shor et al.\\ 2022 \cite{shor2022universal}\end{tabular}  & \begin{tabular}[c]{@{}l@{}}SUPERB benchmark,\\ LibriSpeech (ASR),\\ VoxCeleb1/2 (SV),\\ AMI (SD),\\ CommonVoice (LID),\\ CREMA-D (SER)\end{tabular} & \begin{tabular}[c]{@{}l@{}}SUPERB: 0.9511234,\\ LibriSpeech WER: 2.3/5.61,\\ VoxCeleb1/2 EER: 0.67/0.831,\\ AMI DER: 8.31,\\ CommonVoice LID ACC: 0.9781,\\ CREMA-D SER UAR: 0.726\end{tabular} & \begin{tabular}[c]{@{}l@{}}A stack of convolution-augmented transformer blocks\\ known as Conformers with a total of 600M parameters\\ trained on YT-U dataset using self-supervision.\end{tabular} \\ \hline
\begin{tabular}[c]{@{}l@{}}Xu et al.\\ 2021 \cite{xu2021transformer}\end{tabular}  & \begin{tabular}[c]{@{}l@{}}IEMOCAP,\\ MSP-IMPROV,\\ RAVDESS\end{tabular}  & \begin{tabular}[c]{@{}l@{}}IEMOCAP: WA: 0.661, UA: 0.633;\\ MSP-IMPROV: WA: 0.651, UA: 0.644;\\ RAVDESS: WA: 0.713, UA: 0.712\end{tabular}  & \begin{tabular}[c]{@{}l@{}}A transformer-based end-to-end model that \\consists of a CNN encoder, a transformer encoder, \\a temporal attention layer, and a softmax classifier.\end{tabular} \\ \hline
\begin{tabular}[c]{@{}l@{}}Gao et al.\\ 2022 \cite{gao2022paraformer}\end{tabular} & \begin{tabular}[c]{@{}l@{}}ComParE 2019-2021,\\ IEMOCAP,\\ MSP-IMPROV,\\ RAVDESS,\\ SAVEE,\\ EmoDB,\\ CREMA-D,\\ EMOVO-Corpus,\\ ShEMO-Corpus\end{tabular}  & \begin{tabular}[c]{@{}l@{}}ComParE 2019: UAR: 0.72\\ ComParE 2020: UAR: 0.67\\ ComParE 2021: UAR: 0.65\\ IEMOCAP: F1: 0.71\\ MSP-IMPROV: F1: 0.69\\ RAVDESS: UAR: 0.76\\ SAVEE:  UAR: 0.77\\ EmoDB: UAR: 0.80\\ CREMA-D: UAR: 0.81\\ EMOVO-Corpus: UAR: 0.79\\ ShEMO-Corpus: UAR: 0.78\end{tabular} & \begin{tabular}[c]{@{}l@{}}A hierarchical framework that consists of a \\transformer-based encoder-decoder model \\and a multi-task learning module.\end{tabular} \\ \hline
\begin{tabular}[c]{@{}l@{}}Chen et al.\\ 2023 \cite{chen2023speechformer++}\end{tabular} & \begin{tabular}[c]{@{}l@{}}LibriSpeech \\test-clean/test-other.\end{tabular} & WER: 2.3\%/5.4\%  & \begin{tabular}[c]{@{}l@{}}A parallel transformer that utilizes a continuous \\integrate-and-fire based predictor to predict \\the number of tokens and generate hidden variables.\end{tabular} \\ \hline
\end{tabular}
\label{tab:SpeechParalinguistics-Studies}
\end{table*}

\subsection{Speech Paralinguistics}

Speech paralinguistics is a term that refers to the non-verbal aspects of speech communication, such as tone, pitch, volume, speed, emotion, and accent \cite{schotz2002paralinguistic}. In terms of NLP, speech paralinguistics is an area that aims to analyse and synthesize speech signals with paralinguistic features. These features can convey important information about the speaker’s identity, intention, attitude, and mood, and can enhance the performance and naturalness of various speech applications. In this section, we discuss how transformers can be used for speech paralinguistic tasks, which involve analyzing and synthesizing speech signals with non-verbal features such as emotion, speaker identity, and accent. We focus on a recent paper by Chen et al. (2021) \cite{chen2022wavlm}, which proposes a new pre-trained model called WavLM for full-stack speech processing. WavLM uses a masked speech prediction and a speech-denoising objective to learn universal speech representations from large-scale unlabeled data. WavLM also employs a gated relative position bias mechanism to capture the sequence order of speech signals. The paper shows that WavLM achieves state-of-the-art results on the SUPERB benchmark \cite{yang2021superb} and improves performance on several other speech benchmarks for tasks such as speech emotion recognition (SER), speaker verification (SV), speaker diarization (SD), and speech separation (SS). We review the main contributions of WavLM and compare it with other transformer-based models for speech paralinguistic tasks.

In speech paralinguistics, Xu et al. (2021) proposed a new attention mechanism called local dense synthesizer attention (LDSA) \cite{xu2021transformer}. The mechanism restricts attention scope to a local range around the current frame and eliminates dot products and pairwise interactions to improve the performance of end-to-end speech recognition models while reducing computational complexity. The study also combines LDSA with self-attention to extract both local and global information from speech signals. Shor et al. (2022) proposed a new pre-trained model, Conformer-HuBERT, that combines conformers and HuBERT to learn universal speech representations for paralinguistic tasks such as SER, SV, SD, and SS \cite{shor2022universal}. Conformers are hybrid architectures that integrate CNNs and transformers, while HuBERT is a self-supervised learning framework that learns from large-scale unlabeled data. The study demonstrates that Conformer-HuBERT outperforms existing models on several benchmarks for paralinguistic tasks, achieving state-of-the-art results.

Shor and Venugopalan (2022) \cite{shor2022trillsson} propose a collection of small and performant models called TRILLsson, which are distilled from a large self-supervised model called CAP12 \cite{shor2022universal}. TRILLsson uses knowledge distillation on public data to reduce the size of CAP12 by up to 100x while retaining 90-96 percent of its performance. The paper demonstrates that TRILLsson outperforms previous models on the Non-Semantic Speech (NOSS) benchmark \cite{shor2020towards}. Additionally, the paper releases the TRILLsson models publicly. Another attempt by Chen et al. (2022) \cite{chen2022speechformer} propose a novel framework, SpeechFormer, that incorporates the unique characteristics of speech signals into transformer models. The framework comprises three components: a hierarchical encoder that reduces the input sequence length using convolutional and pooling layers, a local self-attention module that captures dependencies within a fixed window size, and a global self-attention module that captures dependencies across different windows. The paper demonstrates that SpeechFormer achieves competitive results on several speech benchmarks for tasks such as automatic speech recognition (ASR), speaker verification (SV), speaker diarization (SD), and emotion recognition. Another recent paper, SpeechFormer++: by Chen et al. (2023) \cite{chen2023speechformer++} builds on the previous work of SpeechFormer \cite{chen2022speechformer} and incorporates the unique characteristics of speech signals into transformer models. The framework includes a unit encoder that models the intra- and inter-unit information, a merging block that generates features at different granularities based on the hierarchical relationship in speech signals, and a word encoder that integrates word-grained features into each unit encoder. The paper demonstrates that SpeechFormer++ outperforms the standard transformer on various paralinguistic tasks, such as speech emotion recognition (SER), depression classification (DC), and Alzheimer's disease detection (ADD).

Gao et al. (2022) introduced Paraformer, a new model for non-autoregressive end-to-end speech recognition that uses parallel attention and parallel decoder techniques \cite{gao2022paraformer}. Paraformer's encoder-decoder architecture allows each decoder layer to attend to all encoder outputs simultaneously without waiting for previous decoder outputs, and each output token to be predicted independently without depending on previous output tokens. The paper shows that Paraformer outperforms existing non-autoregressive models on several ASR datasets, achieving faster inference speed and higher accuracy. These recent innovations in transformer-based models such as WavLM, Conformer-HuBERT, TRILLsson, SpeechFormer, and Paraformer have shown promising results for speech paralinguistic tasks, paving the way for more natural and efficient speech applications.

\begin{table*}[!ht]
\caption{Recent studies on transformers for \textbf{Speech Enhancement}.}
\centering
\begin{tabular}{|l|l|l|l|}
\hline
\textbf{Author (year)} & \textbf{Datasets}   & \textbf{Performance}  & \textbf{Architecture(s)}   \\ \hline
\begin{tabular}[c]{@{}l@{}}Yu et al.\\ 2022 \cite{yu2022setransformer}\end{tabular} & \begin{tabular}[c]{@{}l@{}}VCTK,\\ CHiME-3\end{tabular} & \begin{tabular}[c]{@{}l@{}}PESQ: 2.97 ± 0.01,\\ STOI: 0.94 ± 0.00,\\ SI-SNRi: 16.9 ± 0.1 dB\end{tabular}  & \begin{tabular}[c]{@{}l@{}}Encoder-LSTM-Multi-head attention-\\ Decoder\end{tabular}   \\ \hline
\begin{tabular}[c]{@{}l@{}}Kim et al.\\ 2020 \cite{kim2020t}\end{tabular} & VCTK    & \begin{tabular}[c]{@{}l@{}}PESQ: 3.06 ± 0.01,\\ STOI: 0.95 ± 0.00,\\ SI-SNRi: 17.8 ± 0.1 dB\end{tabular}  & \begin{tabular}[c]{@{}l@{}}Encoder-Self-attention with Gaussian \\ weights-Decoder\end{tabular}    \\ \hline
\begin{tabular}[c]{@{}l@{}}Wang et al.\\ 2021 \cite{wang2021tstnn}\end{tabular} & \begin{tabular}[c]{@{}l@{}}VCTK,\\ DNSCL,\\ DNSCL-R\end{tabular}    & \begin{tabular}[c]{@{}l@{}}PESQ:\\ Voice Bank + DEMAND: 3.08,\\ DNSCL: 3.02,\\ DNSCL-R: 2.93;\\ STOI:\\ Voice Bank + DEMAND: 0.95,\\ DNSCL: 0.94,\\ DNSCL-R: 0.92\end{tabular}  & \begin{tabular}[c]{@{}l@{}}Encoder-Two-stage transformer module-\\ Masking module-Decoder\end{tabular}   \\ \hline
\begin{tabular}[c]{@{}l@{}}Subakan et al.\\ 2021 \cite{subakan2021attention}\end{tabular} & \begin{tabular}[c]{@{}l@{}}WSJ0-2mix,\\ WSJ0-3mix\end{tabular}  & \begin{tabular}[c]{@{}l@{}}WSJ0-2mix:\\ SDR: 20.8 dB, SI-SNR: 21.9 dB;\\ WSJ0-3mix:\\ SDR: 17.6 dB, SI-SNR: 18.7 dB\end{tabular}  & \begin{tabular}[c]{@{}l@{}}Multi-scale transformer with multi-head \\ attention and feed-forward layers\end{tabular} \\ \hline
\begin{tabular}[c]{@{}l@{}}Zhang et al.\\ 2022 \cite{zhang2022cross}\end{tabular}   &\begin{tabular}[c]{@{}l@{}}LibriSpeech,\\ LibriMix,\\ WHAMR,\\ WHAM\end{tabular} &  \begin{tabular}[c]{@{}l@{}}LibriMix:\\ PESQ: 3.25, STOI: 0.95, ESTOI: 0.93;\\ WHAMR:\\ PESQ: 2.98, STOI: 0.94, ESTOI: 0.91;\\ WHAM\\ PESQ: 3.04, STOI: 0.95, ESTOI: 0.92;\end{tabular} & \begin{tabular}[c]{@{}l@{}}Streaming transformer with cross-attention \\ between encoder and decoder layers\end{tabular}   \\ \hline
\begin{tabular}[c]{@{}l@{}}Zhao et al.\\ 2021 \cite{zhao2021multi}\end{tabular} &  \begin{tabular}[c]{@{}l@{}}WSJ0-2mix,\\ WSJ0-3mix\end{tabular} & \begin{tabular}[c]{@{}l@{}}WSJ0-2mix:\\ SDR: 20.6 dB, SI-SNR: 21.7 dB;\\ WSJ0-3mix9:\\ SDR: 17.5 dB, SI-SNR: 18.6 dB\end{tabular} & \begin{tabular}[c]{@{}l@{}}Multi-scale group transformer with dense-\\ fusion or light-fusion and time-domain\\ audio separation network (TasNet)\end{tabular} \\ \hline
\begin{tabular}[c]{@{}l@{}}Jiang et al.\\ 2023 \cite{jiang2023low}\end{tabular} & LibriSpeech   & \begin{tabular}[c]{@{}l@{}}PESQ: 3.25,\\ STOI: 0.95,\\ ESTOI: 0.93,\\ SI-SNRi: 19.1 dB\end{tabular} & \begin{tabular}[c]{@{}l@{}}Hierarchical frame-level Swin\\ Transformer with adaptive windowing and\\ convolutional layers\end{tabular} \\ \hline
\end{tabular}
\label{tab:SpeechEnhancement-Studies}
\end{table*}

\subsection{Speech Enhancement and Separation}
The area of speech enhancement involves the application of various algorithms for enhancing the quality of speech. 
Speech enhancement (SE) aims to isolate the speech of a targeted user from a group of others. Previously, neural networks have been employed in an attempt to achieve this goal. One such implementation utilized an audio embedding network to extract the audio embedding of different speakers, which was then utilized in a spectrogram masking network to produce an output with masks \cite{wang2018voicefilter}. This approach yielded a more efficient and faster model as the embedding for each speaker was computed in advance. The features extracted were subsequently employed in the PSE network for enhancing the speech signals of specific users, enabling the segregation of separate networks and allowing for further individual improvements \cite{wang2018voicefilter,wang2020voicefilter}. At inference, the speaker embedding is concatenated with the intermediate features of the PSE network for conditionality purposes. In another approach, audio signal embedding vectors representing the desired speaker were utilized to improve noise and echo cancellation and speech enhancement \cite{o2021conformer}. Another model, known as the Sound-Filter model, employed unlabeled data for speech enhancement. This wave-to-wave convolutional neural network was trained using mixtures generated from a collection of unlabeled audio recordings. It was assumed that speech was from a single source for the entire duration of a clip, based on the use of short intervals of audio signals for the same type of sound. This problem was approached as a one-shot learning challenge, resulting in models with conditioning encoder clusters that mapped acoustically similar sounds together in the embedding space \cite{gfeller2021one}. Additionally, activation functions were learned to personalise the output \cite{ramos2022conditioning}.

The application of speech separation is a crucial aspect of speech processing, and Recurrent Neural Networks (RNNs) were predominantly utilized for this purpose since their introduction. However, the trend has shifted towards the usage of Transformers as they enable parallel computation through the attention mechanism. The sequence-modeling capabilities of Transformers have the potential to enhance speech separation. An example of this approach is the proposed SepFormer model in \cite{subakan2021attention}. The authors leveraged the parallel computation benefits of Transformers in the implementation of their model. Furthermore, the utilization of Transformers in speech separation has been observed in other studies as well, such as the work of Zhang et al. \cite{zhang2021transmask}. The issue of increased computational complexity with longer sequences of sentences in the field of speech separation has been addressed through the utilization of the multi-scale group transformer (MSTG) approach. As reported by Zhao et al. \cite{zhao2021multi}, this approach leverages the self-attention capabilities of transformers and incorporates multi-scale fusion to capture long-term dependencies, thereby reducing computation complexity. The size of state-of-the-art models for speech separation tasks, however, often reaches hundreds of gigabytes, presenting a common challenge in the field. 

To mitigate this challenge, various approaches have been employed including Knowledge Distillation \cite{hinton2015distilling}. The Teacher-Student model \cite{chen2022ultra} has been utilized in an approach aimed at reducing the size and complexity of models for speech separation. Another approach reported in \cite{subakan2022resource} achieved this through the utilization of non-overlapping blocks in latent space and compact latent summaries calculated for each chunk. The authors developed the RE-SepFormer and achieved performance on par with existing state-of-the-art models. Another innovative approach, Tiny-Sepformer \cite{luo2022tiny}, uses a time-domain transformer neural network and achieved this by splitting Convolution Attention (CA) blocks and implementing parameter sharing within CA blocks.

\subsection{Spoken Dialogue Systems}
\begin{figure}[!t]
\centering
\includegraphics[width=0.49\textwidth]{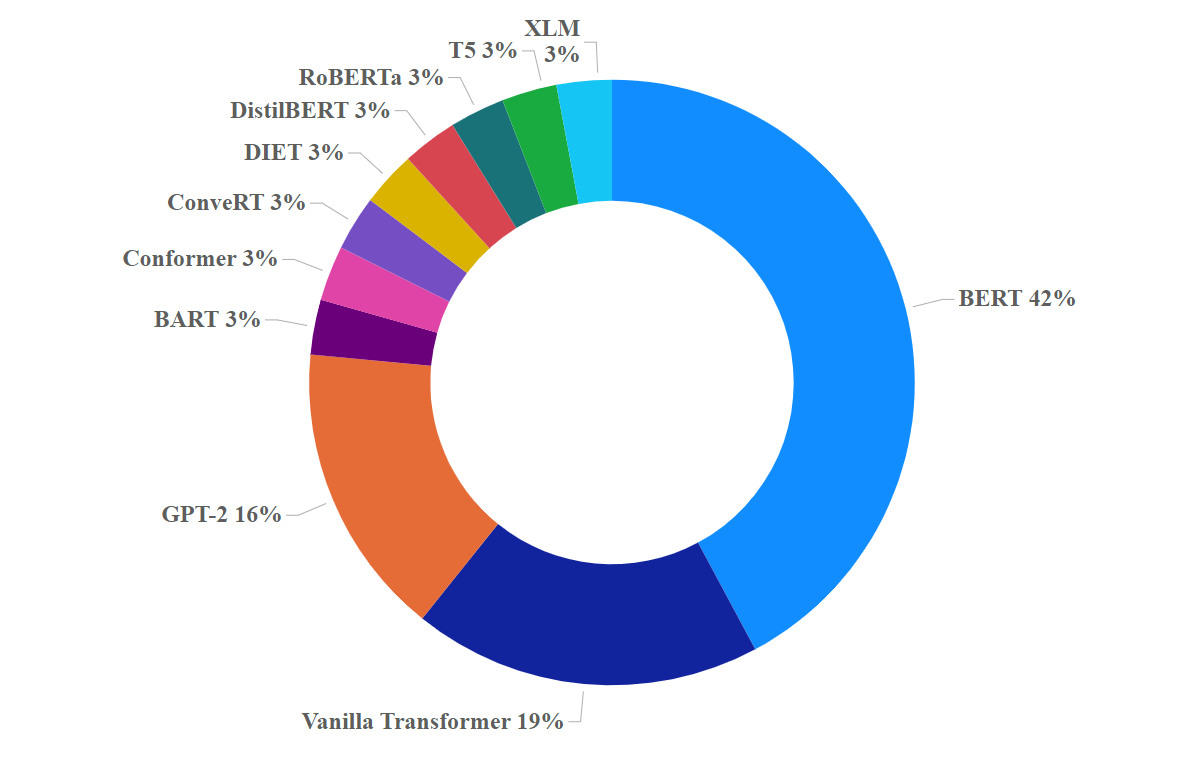}
\caption{Transformer-based architectures in Spoken Dialogue Systems.}
\label{fig:sds-transformers}
\end{figure}

Table~\ref{table:sds-transformers} shows a list of related works on spoken dialogue systems using Transformer networks. But it should be noted that most of those neural architectures have been originally applied to text-based language processing tasks, not to speech data with some exceptions \cite{GulatiQCPZYHWZW20,DIET}. From the popularity chart in Fig.~\ref{fig:sds-transformers}, it can be noted that the most popular neural architecture is BERT (Bidirectional Encoder Transformer) \cite{DevlinCLT19} and the second most popular choice of architecture is either the original/vanilla transformer architecture \cite{vaswani2017attention} or GPT-2 (Generative Pre-Trained Transformer) \cite{radford2018improving}. Whilst BERT and GPT-2 are generalizations of the vanilla transformer networks, BERT uses encoder blocks (no decoder blocks) and bidirectional representations whilst GPT-2 uses decoder blocks (no encoder blocks) and left-to-right representations. Other generalizations of Transformers include the following: XLM (Cross-lingual Language Model) \cite{ConneauL19} to benefit from data in different languages; DistillBERT (Distilled version of BERT) \cite{DistilBERT} to train more compact models via knowledge distillation; ConvERT \cite{HendersonCMSWV20} to perform faster training via pre-trained response selection; BART (Bidirectional Auto-Regressive Transformers) \cite{LewisLGGMLSZ20} to pre-train sequence-to-sequence models via a denoising autoencoder (by predicting outputs without noise from noisy inputs); T5 (Text-to-Text Transfer Transformer) \cite{RaffelSRLNMZLL20} to learn from data in multiple language tasks by converting it to text-to-text format and then carrying out transfer learning to a specific language task; Conformer (Convolution-augmented Transformer) \cite{GulatiQCPZYHWZW20} to bring the advantages of Transformer and Convolutional neural nets into a single architecture suitable for audio sequences; DIET (Dual Intent and Entity Transformer) \cite{DIET} to perform language understanding of intents and entities in utterances without pretraining; GPT-3 (large Generative Pre-Trained Transformer) \cite{BrownMRSKDNSSAA20} to learn large language models without the need of fine-tuning; and RoBERTa (Robustly optimised BERT approach) \cite{LiuLSZ21} to learn language models with an improved methodology over BERT. Some other related architectures have been proposed and trained with text instead of speech data \cite{TransferTranfo,BudzianowskiV19,LiuWLXF20,ZhangSGCBGGLD20,HODChatbot,MadottoCW0LLF20,LinLHNGHFG21,RohmatillahC21,EppsEtAl,Sun0BRRCR21,RollerDGJWLXOSB21,WuJ22,AbroARUMQ22,GODEL,Jang0K22,MHC}.

\begin{figure}[!t]
\centering
\includegraphics[width=0.49\textwidth]{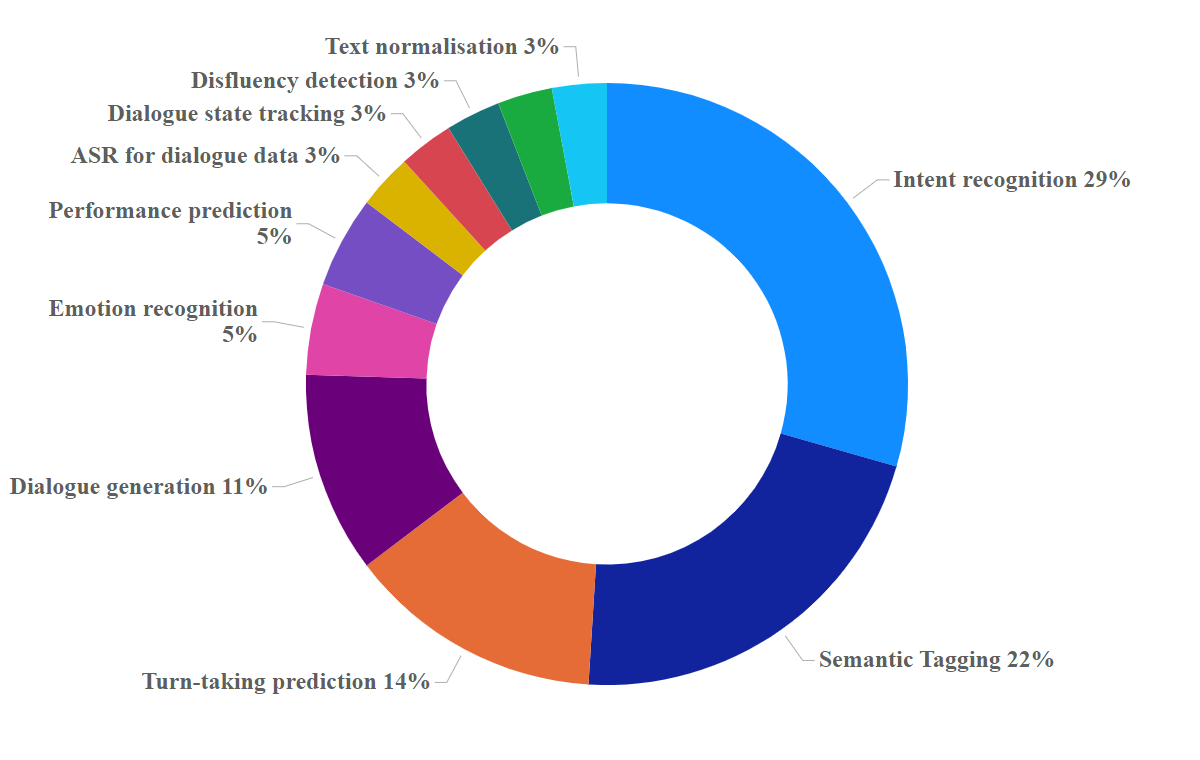}
\caption{Tasks of Transformer-based Spoken Dialogue Systems.}
\label{fig:sds-tasks}
\end{figure}

From Fig.~\ref{fig:sds-tasks}, it can be noted that Transformer networks have been mostly applied to language understanding tasks (52\% of related works) including intent recognition and semantic tagging (a.k.a. slot filling). They are followed by turn-taking prediction and dialogue generation\footnote{Dialogue generation in this paper refers to generating a system response either word by word or by selecting a sentence from a pool of candidates). } (25\% of related works). Other less popular application areas (23\% of related works) include emotion recognition, performance prediction (of language understanding or user satisfaction), dialogue state tracking, punctuation prediction, disfluency detection, text normalization, and speech recognition for conversational data. The full set of tasks gives us an idea of the range of skills involved in a spoken dialogue system. 

\begin{table*}[!ht]
\scriptsize
\caption{Representative list of publications in Transformer-based \textbf{Spoken Dialogue Systems} in the period 2019-2023.}
\begin{center}
\begin{tabular}{|l|l|l|l|}
\hline
Author (year) & Architecture(s) & Task(s) & Dataset(s)\\
\hline
\hline
Takatsu et al. 2019 \cite{TakatsuYMHFK19} & BERT & Intent recognition & Majority Vote\\
\hline
Chen et al. 2019 \cite{SlotCarryOver}  & Vanilla Transformer & Dialog state tracking & DSTC2\\
\hline
Liu et al. 2019 \cite{BERT_BiLSTM} & BERT, BiLSTM & Intent recognition & CamRest\\
\hline
Korpusik et al. 2019 \cite{KorpusikLG19} & BERT & Semantic tagging & ATIS, Restaurants\\
\hline
Zhao H. Wang. 2019 \cite{ZhangW19} & Vanilla Transformer, CRF & Intent recognition, semantic tagging & ATIS, SNIPS\\
\hline
Quian et al. 2020 \cite{QianSZ20} & GPT-2 & Semantic tagging & DSTC2\\
\hline
Hong et al. 2020 \cite{HongKK20} & BERT & Semantic tagging & CamRest\\
\hline
Ekstedt et al. 2020 \cite{EkstedtS20} & GPT-2 & Turn-taking prediction & Maptask, Switchboard\\
\hline
Gopalakrishnan et al. 2020 \cite{0001HWLH20} & GPT-2 & Dialogue generation: writen vs. spoken & Topical-chat\\
\hline
Chen et al. 2020 \cite{ChenCLW20} & Vanilla/CT Transformer & Punctuation/disfluency prediction & IWSLT2011, in-house Chinese data\\
\hline
Hori et al. 2020 \cite{HoriMHR20} & Vanilla Transformer & ASR specialized for conversational data & Switchboard, HKUST\\
\hline
Lian et al. 2021 \cite{Lian0T21} & Vanilla Transformer & Emotion recognition in dialogue & IEMOCAP, MELD\\
\hline
Andreas et al. 2021 \cite{ComparativeStudy} & BART, GPT, BERT & Dialogue generation: model comparison & CamRest\\
\hline
Chapuis et al. 2021 \cite{ChapuisCMLC20} & BERT, hierarchical arch. & Emotion recognition, dialogue act recog. & SILICONE\\
\hline
Lai et al. 2021 \cite{TextNorm21} & BERT, RoBERTa & (Inverse) Text normalization & Google TN dataset\\
\hline
Kim et al. 2021 \cite{KimKSL21} & BERT & Intent recognition/classification & FSC, SNIPS, SmartLights\\
\hline
Lopez-Zorrilla et al. 2021 \cite{Lopez-ZorrillaT21} & GPT-2, RL & Dialogue generation: policy learning & DSTC2\\
\hline
Bechet et al. 2021 \cite{BechetRHAMD21} & BERT & SLU performance prediction & ATIS, MEDIA, SNIPS, M2M\\
\hline
Okur et al. 2022 \cite{OkurEtAl} & BERT, ConveRT, DIET & Intent recognition & Math-Game\\
\hline
Sakuma et al. 2022 \cite{SakumaFK22} & Conformer & End-of-utterance prediction & Japanese Restaurant Data\\
\hline
Lin et al. 2022 \cite{LinWHSSL22} & Vanilla Transformer & User-state/barge-in/backchannel detection & Chinese dialogues\\
\hline
Ahmed 2022 \cite{AhmedPhDThesis} & BERT & Intent recognition, semantic tagging & ATIS, SNIPS\\
\hline
Bekal et al. 2022 \cite{BekalEtAl} & BERT, HuBERT & Turn-taking prediction: barge-in detection & 12k chat-bot prompts\\
\hline
Dong et al. 2022 \cite{DongFZLW22} & BERT & Intent recognition/classification & FSC, SmartLights\\
\hline
Svec et al. 2022 \cite{SvecFBL22} & T5 & Intent recognition/classification & HHTT, TIA (in Czech)\\
\hline
Shen et al. 2022 \cite{ShenHZZX22} & Vanilla Transformer, TBM & User satisfaction prediction & Industry data: 18M/32M/5M samples\\
\hline
Yang et al. 2022 \cite{YangWZFCH22} & Vanilla Transformer & Turn-taking prediction & Commercial phone-dialogues\\
\hline
Sunder et al. 2022 \cite{SunderTKGKF22} & BERT & Dialogue act recognition & HarperValleyBank\\
\hline
Lopez-Zorrilla et al. 2022 \cite{Lopez-ZorrillaT23} & GPT-2, RL & Dialogue generation: policy learning & DSTC2\\
\hline
Firdaus et al. 2023 \cite{FirdausEC23} & mBERT, RoBERTa, XML & Intent recognition, semantic tagging & ATIS, TRAINS, SNIPs, FRAMES\\
\hline
Mei et al. 2023 \cite{MeiWTDH23} & BERT & Intent recognition, semantic tagging & ATIS, SNIPS\\
\hline
\end{tabular}

\label{table:sds-transformers}
\end{center}
\end{table*}

Regarding the performance of Transformers, BERT models have been shown to outperform their predecessor recurrent/convolutional neural nets in language understanding tasks (semantic tagging and intent recognition) \cite{BERT_BiLSTM,KorpusikLG19,HongKK20,FirdausEC23,MeiWTDH23}, text normalization \cite{TextNorm21}, performance prediction of language understanding \cite{BechetRHAMD21}, and turn-taking prediction \cite{BekalEtAl} – though not always clearly and sometimes with only small differences. Using the popular stages of pre-training and fine-tuning has been reported to achieve better performance over using only one stage avoiding pre-training \cite{TakatsuYMHFK19,KimKSL21,DongFZLW22}. When extended in a hierarchical way to model word sequences at the low level, utterances at the mid-level, and dialogues at the high level, a hierarchical BERT is able to outperform its single model counterpart \cite{ChapuisCMLC20}. Hierarchical models are also trained with fewer parameters than their counterpart non-hierarchical models \cite{SunderTKGKF22}. Similarly, when combining specialised architectures such as DIET and ConVERT has shown improved results over using standard Transformers with pre-trained BERT embeddings \cite{OkurEtAl}. Even when combining more standard architectures such as BERT with biLSTMs (bidirectional Long-Short Term Memory Networks) has shown improved results over BERT-like baselines \cite{AhmedPhDThesis}. Moreover, BERT models have also shown to benefit from using data augmentation to obtain further gains over only pre-training and fine-tuning \cite{KimKSL21}. Furthermore, BERT models can also benefit from combining multiple modalities (such as audio and text) and using a cross-modal contrastive loss over using a single modality \cite{DongFZLW22}. 

Similarly, GPT-2 models have shown to outperform their predecessor recurrent neural nets in turn-taking prediction \cite{EkstedtS20}. GPT-2 models have also shown promising results by training a semantic tagger jointly with a speech recognizer and outperforming independent models \cite{QianSZ20}. When training GPT-2-based models on text data and testing on noisy data derived from speech recognition, the selected responses may be out of context—suggesting the need for training Transformer-based spoken dialogue systems using noisy data \cite{0001HWLH20}. The latter is supported by the work of \cite{Lopez-ZorrillaT21,Lopez-ZorrillaT23}, who extended GPT-2 pre-trained language models with audio embeddings in order to train models with improved performance over using only text-based representations. When extending GPT-2 models (among other Transformer-based architectures) with knowledge embeddings (KE) to port a knowledge base into the model parameters \cite{MadottoCW0LLF20}, the KE-enhanced models outperform the KE-unaware models \cite{ComparativeStudy}. In a similar vein, vanilla transformers have shown to outperform their predecessor recurrent neural nets (RNN) in tasks such as dialogue state tracking \cite{SlotCarryOver}, intent recognition and semantic tagging \cite{ZhangW19}, punctuation prediction and disfluency detection \cite{ChenCLW20}, speech recognition for conversational data \cite{HoriMHR20}, emotion recognition \cite{Lian0T21}, user satisfaction prediction \cite{ShenHZZX22}, and turn-taking prediction \cite{YangWZFCH22}. Whilst transformers have mostly reported positive improvements over RNN baselines across different spoken language processing tasks, \cite{LinWHSSL22} found no significant differences between them in the task of turn-taking prediction using a dataset of 10K labeled instances. 

Some recent models have shown further improvements over vanilla, BERT, or GPT-2 Transformers. The following are some examples. \cite{ComparativeStudy} have reported BART and T5 models to outperform vanilla and GPT-2 Transformers in the task of dialogue generation. \cite{OkurEtAl} have shown that models combining DIET and ConVERT are able to outperform vanilla transformers with BERT embeddings in the task of intent recognition. \cite{FirdausEC23} have found that multilingual multi-task BERT models are able to outperform strong baselines such as RoBERTa and XLM in the tasks of intent recognition and semantic tagging.

Other recent works in Transformer-based dialogue systems but using text data are also mostly based on BERT-based architectures \cite{LiuWLXF20,RohmatillahC21,EppsEtAl,Sun0BRRCR21,RollerDGJWLXOSB21,WuJ22,AbroARUMQ22} or GPT-based architectures \cite{TransferTranfo,BudzianowskiV19,ZhangSGCBGGLD20,MadottoCW0LLF20,GODEL,Jang0K22,MHC}. Notable large-scale efforts include GODEL \cite{GODEL} and InstructGPT \cite{InstructGPT}. GODEL (Grounded Open Dialogue Language Model) is framed as suitable for open-ended goal-directed dialogue, in its base and large versions containing 220M and 770M parameters, has been shown to outperform BART, T5, DialoGPT and GPT-3 across different benchmarks. Those results are based on automatic and human evaluations. InstructGPT, a fine-tuned GPT-3 model using reinforcement learning with human feedback, outperforms GPT-3 baselines that do not learn from human feedback. The latter results in a more compact model of 1.3 billion parameters being preferred over a larger model of 175 billion parameters. InstructGPT is the neural architecture of the widely known dialogue system ChatGPT\footnote{\url{https://openai.com/blog/chatgpt}}. 

Although previous works have shown remarkable progress in spoken and text-based language processing using Transformer networks, however, it is unclear which is the best neural architecture for a particular task or across tasks. Some works include a few baselines and some others have different baselines, and there is a wide variety of datasets being used---due to the number of tasks involved in dialogue systems. While performing such comparisons requires significant resources in terms of data and computing power, they would be highly valuable for establishing a clearer understanding of the current state-of-the-art in the field. Efforts such as GLUE in NLP \cite{WangSMHLB19} are needed in spoken dialogue systems. Nonetheless, the battle so far seems to be between BERT and GPT---and novel architectures performing even better remain to be discovered. While there have been some notable successes in conversational AI, the average citizen has yet to fully benefit from them. Many services, whether accessed over the phone or through the web, still rely on rudimentary methods, such as requiring people to fill out long web-based forms or endure long wait times to speak with customer representatives. However, this is likely to change as technologies such as Transformer-based text and spoken dialogue systems become more accessible and easier to deploy. Additionally, given that experimental results in this field continue to show room for improvement in the correctness and safety of generated responses, there is a need for more effective methods to be developed.




\begin{table*}[!ht]
\centering
\caption{Recent studies on transformers for \textbf{Multi-Modal Applications}.}
\scriptsize
\begin{tabular}{|l|l|l|l|}
\hline
\textbf{Author (year)}  & \textbf{Datasets}    & \textbf{Performance}  & \textbf{Architecture(s)}    \\ \hline
\begin{tabular}[c]{@{}l@{}}Chuang et al.\\ 2019 \cite{chuang2019speechbert}\end{tabular} & \begin{tabular}[c]{@{}l@{}}LibriSpeech,\\ TIMIT,\\ SQuAD v2.01,\\ SWBD-Fisher\end{tabular}   & \begin{tabular}[c]{@{}l@{}}LibriSpeech: \\ WER 9.8\% (dev-clean),\\ 23.4\% (dev-other);\\ TIMIT: PER 14.7\% (test);\\ SQuAD v2.0:\\ F1 score 76.6\%,\\ EM score 69.8\%;\\ SWBD-Fisher: WER 10.9\%\end{tabular}  & \begin{tabular}[c]{@{}l@{}}BERT-base model with a speech encoder consisting of two \\convolutional layers and four self-attention layers  \end{tabular}   \\ \hline
\begin{tabular}[c]{@{}l@{}}Song et al.\\ 2019 \cite{song2019speech}\end{tabular}   & \begin{tabular}[c]{@{}l@{}}TIMIT; \\ WSJ\end{tabular}  & \begin{tabular}[c]{@{}l@{}}TIMIT: PER 12.5\% (test);\\ WSJ: \\ WER 11.4\% (dev93), 10.4\% (eval92)\end{tabular} & \begin{tabular}[c]{@{}l@{}}XLNet-base model with a speech encoder consisting of two \\convolutional layers and six self-attention layers.\end{tabular}  \\ \hline
\begin{tabular}[c]{@{}l@{}}Ao et al.\\ 2021 \cite{ao2021speecht5}\end{tabular} & \begin{tabular}[c]{@{}l@{}}LibriSpeech;\\ LibriTTS;\\ Common Voice;\\ TIMIT;\\ WSJ;\\ LJSpeech;\\ VCTK\end{tabular}  & \begin{tabular}[c]{@{}l@{}}LibriSpeech: WER 4.0\% (test-clean),\\ 10.9\% (test-other)1;\\ LibriTTS: MOS 4.011;\\ Common Voice: WER 6.8\% (en);\\ TIMIT: PER 9.7\% (test);\\ WSJ: WER 6.0\% (dev93),\\ 5.7\% (eval92)1;\\ LJSpeech: MOS 4.021; VCTK: MOS 3.97\end{tabular} & \begin{tabular}[c]{@{}l@{}}T5-base model with a speech encoder consisting of \\two convolutional layers and six self-attention layers.\end{tabular}   \\ \hline
\begin{tabular}[c]{@{}l@{}}Arjmand et al.\\ 2021 \cite{arjmand2021teasel}\end{tabular}   & CMU-MOSI & \begin{tabular}[c]{@{}l@{}}Accuracy score of 76\%,\\ F-score of 0.75,\\ MAE score of 0.94 on test set\end{tabular}    & \begin{tabular}[c]{@{}l@{}}A pre-trained language model such as BERT or RoBERTa\\ with a speech prefix consisting of two \\convolutional layers and one self-attention layer.\end{tabular}  \\ \hline
\begin{tabular}[c]{@{}l@{}}Sant et al.\\ 2022 \cite{sant2022multiformer}\end{tabular}    & \begin{tabular}[c]{@{}l@{}}MuST-C,\\ CoVoST v2\end{tabular}    & \begin{tabular}[c]{@{}l@{}}MuST-C: BLEU 25.9 (en-de),\\ 28.8 (en-es), 29.1 (en-fr),\\ 20.3 (en-it), 21.0 (en-nl),\\ 22.8 (en-pt), 18.9 (en-ro);\\ CoVoST v2:\\ BLEU 24.1 (es-en), 23.3 (fr-en)\end{tabular}   & \begin{tabular}[c]{@{}l@{}}A Transformer model with a head-configurable self-\\attention module that allows the use of different attention \\mechanisms in each head.  \end{tabular} \\ \hline
\begin{tabular}[c]{@{}l@{}}Lin et al.\\ 2022 \cite{lin2022compressing}\end{tabular}  & \begin{tabular}[c]{@{}l@{}}LibriSpeech,\\VoxCeleb,\\VoxCeleb\end{tabular}
& \begin{tabular}[c]{@{}l@{}}LibriSpeech:\\ WER 3.0\% (test-clean),\\ 7.6\% (test-other);\\ VoxCeleb1: EER 1.67\%;\\ VoxCeleb2: EER 2.12\%\end{tabular}   & \begin{tabular}[c]{@{}l@{}}A simplified version of HuBERT with a \\convolutional encoder and a Transformer decoder \end{tabular}   \\ \hline
\begin{tabular}[c]{@{}l@{}}Chung et al.\\ 2018 \cite{chung2018speech2vec}\end{tabular}   & \begin{tabular}[c]{@{}l@{}}TIMIT,\\ Buckeye Corpus\end{tabular}  & \begin{tabular}[c]{@{}l@{}}TIMIT:\\ Accuracy 0.81 (skip-gram),\\ 0.80 (cbow);\\ Buckeye Corpus:\\ Accuracy 0.83 (skip-gram),\\ 0.82 (cbow)\end{tabular} & \begin{tabular}[c]{@{}l@{}}A sequence-to-sequence model with an RNN encoder\\ and decoder that learns fixed-length vector representations \\of speech segments.\end{tabular}   \\ \hline
\begin{tabular}[c]{@{}l@{}}Li et al.\\ 2019 \cite{li2019neural}\end{tabular} & \begin{tabular}[c]{@{}l@{}}LJSpeech,\\ Blizzard2012,\\ Blizzard2011\end{tabular} & \begin{tabular}[c]{@{}l@{}}LJSpeech:\\ MOS 4.13 ± 0.08, MAE:0.13110;\\ Blizzard2012:\\ MOS 4.03 ± 0.07, MAE:0.13610;\\ Blizzard2011:\\ MOS 4.01 ± 0.07, MAE:0.138\end{tabular}  & \begin{tabular}[c]{@{}l@{}}A non-autoregressive Transformer model with a multi-\\head self-attention network and a feed-forward network for \\both encoder and decoder.\end{tabular} \\ \hline
\end{tabular}
\label{tab:Multimodal-Studies}
\end{table*}

\subsection{Multi-Modal Applications}
In human-computer interaction, a modality refers to the representation of a human sense using an individual channel of sensory input/output. The modalities in computing encompass vision, audition, reaction, gustation, and olfaction, among others. Multi-modal learning encompasses the use of multiple modalities in conjunction to solve various real-world applications, in much the same way as humans use their multiple senses to complete tasks. For instance, an image of a road sign board can provide an understanding of the type of sign being displayed, while the text on the board adds further context. To mimic this process in computers, data must be solved as separate problems, such as NLP and computer vision. Multi-Modal Learning (MML) provides a general approach to constructing robust models utilizing information from multi-modal data \cite{baltrusaitis2019multimodal}. In order to achieve generalized models in the real world, it is necessary to have excellent models of certain modalities and to use them in conjunction with one another, as some modalities are interdependent in context and deeper understanding, especially in speech processing and NLP-related problems and tasks \cite{zadeh2017tensor}. Transformers have demonstrated promise in achieving good generalization for multi-modal applications, as evidenced by the utilization of multi-model transformers in building AI models for classification \cite{lee2020parameter,nagrani2021attention}, segmentation \cite{strudel2021segmenter}, and cross-modal retrieval \cite{kim2021vilt}.


Recently, there has been a shift towards developing innovative fusion models and architectures. In 2019, the Multimodal Transformer (MulT) was proposed \cite{tsai2019multimodal} to address the issues of long-range dependencies across modalities and non-alignment of data with different sampling rates through the utilization of the attention mechanism in transformers. Another approach was presented in \cite{sun2019videobert} where the authors aimed to learn low-level representations by utilizing both visual and linguistic modalities to generate high-level features without explicit supervision. This model, built on top of BERT, utilized bidirectional joint distributions over sequences of visual and linguistic tokens, derived from vector quantization of video data and speech recognition outputs respectively, producing state-of-the-art results. A similar approach was presented in \cite{huang2021unifying} with the proposal of a unified image-and-text generative framework based on a single multi-modal model to jointly study bi-directional tasks. The use of an Encoder architecture for learning generalized representations has gained significant popularity in recent times. A notable effort was made by Nagrani et al. \cite{nagrani2021attention} to implement this architecture for multi-modal applications, utilizing fusion bottlenecks for the integration of data from various modalities at multiple layers. This fusion process enabled the bottleneck layers to learn more comprehensive representations of the data, leading to improved performance and reduced computational expenses.

Self-supervised learning has also been utilized as a solution to multi-modal problems, with a majority of such work being focused on video and image applications. Zellers et al. \cite{zellers2021merlot} introduced a model, MERLOT, which leveraged self-supervised learning to acquire multi-modal script knowledge through observation of videos transcribed with speech. The model was pre-trained with both frame-level and video-level targets, allowing it to contextualize the data globally. This approach has gained widespread popularity in the video domain due to the abundance of video data available on the internet. Gabeur et al. \cite{gabeur2020multi} proposed a multi-modal transformer, which jointly encodes different modalities in video and facilitates their mutual attention, enabling the encoding and mapping of temporal information. A novel modification to the Encoder architecture was presented in the work of Chen et al. \cite{chen2020uniter}. The authors proposed a joint random masking technique applied to two modalities and utilized conditional masking for pre-training tasks, resulting in the creation of the UNITER model. Another self-supervised approach was introduced in the study by Akbari et al. \cite{akbari2021vatt}, where their model, VATT, took raw input and extracted multi-modal representations through a tokenizer layer, embedding layer, and transformer. This approach leveraged the attention mechanism of transformers to learn the representations of data, producing a model that was more robust for visual and language tasks than prior modality-specific models. The popularity of adversarial learning in creating robust models was also applied, as demonstrated in the work of Li et al. \cite{li2020closer}, where an adversarial learning model was implemented on noise input to produce the improved MANGO model on top of UNITER. This approach achieved state-of-the-art results in terms of robustness benchmarks.

The application of Transformers in multi-modal systems is made feasible by their non-recurrent architecture, which enables sequential modeling. The attention mechanism of Transformers allows for learning across a sequence. The Factorised Multi-modal Transformer (FMT) \cite{zadeh2019factorized} is a new model that makes use of Transformers in multi-modal applications and offers an improvement over existing state-of-the-art models through asynchronous modeling of both intra-modal and inter-modal dynamics. The input in the architecture is first passed through an embedding layer, followed by Multi-modal Transformer Layers (MTL), where each MTL comprises multiple Factorised Multimodal Self-attentions (FMS) that factor in inter-modal and intra-modal aspects of the multi-modal input. The authors of FMT conducted evaluations of its zero-shot task performance and examined if the model learns general representations from pre-trained models. Moreover, they demonstrated that a reduction in the model's losses does not always translate to expected performance gains in multi-modal Transformers. Research has shown that multi-modal Transformer models outperform deeper models with modality-specific attention mechanisms when compared with modality-specific models \cite{hendricks2021decoupling}.





\section{Challenges and Future Work}
\label{sec:challenges}

\subsection{Training Challenges}
Transformers have been proven very effective in speech-related tasks as presented in Section \ref{sec:applications}. However, transformers' training is complex and requires non-trivial efforts regarding carefully designing cutting-edge optimizers and learning rate schedulers \cite{liu2020understanding}. The challenge in terms of applying self-attention to speech recognition is that individual speech frames are not like lexical units such as words. Speech frames do not convey distinct meanings or perform unique functions, which makes it hard for the self-attention mechanism to compute proper attentive weights on speech frames. Considering that adjacent speech frames could form a chunk to represent more meaningful units like phonemes, some sort of pre-processing mechanisms such as convolutions to capture an embedding for a group of nearby speech frames would be helpful for self-attention. Transformers were originally proposed for machine translation, where sequence lengths are short in contrast to speech technology. For instance, sequence lengths in SER are larger and contain a few thousand frames. Self-attention encoders in transformers have quadratic computational complexity and computation of self-attention between all the pairs of frames is expensive to compute. In addition, speech sequences are less informationally dense compared to the word sequences in textual data. Therefore, researchers exploit tricks including time-restricted self-attention \cite{povey2018time}, truncated self-attention \cite{yeh2019transformer}, down-sampling \cite{sperber2018self}, sub-sampling \cite{salazar2019self} and pooling \cite{dong2019self} are being used as transformers in speech technology to tackle sequence length problems.   Positional encoding is another main component in transformers to include a piece of positional information about each word about its position in the sentence. This helps the transformers to capture longer dependencies in sequential data. The original paper utilized sinusoidal position encoding, which can hurt performance in speech-based systems due to longer sequences \cite{bie2019simplified} and generalize poorly in certain conditions \cite{likhomanenko2021cape}. Different approaches \cite{mohamed2019transformers,likhomanenko2021cape} have been explored to address this issue. However, these approaches are exploited in ASR and further research is required in other speech-related domains.

\subsection{Computational Cost and Efficiency}
Recently, transformer-based end-to-end models have achieved great success in many speech-related areas. However, compared to LSTM models, the heavy computational cost of the transformer during inference is a key issue to prevent their applications \cite{lu2020exploring}. The computational cost of the Transformer Transducer grows significantly with respect to the input sequence length, which obstacles the practical use of T-T. Recently conformer Transducer (C-T) \cite{gulati2020conformer} was proposed to further improve T-T, but it is not streamable because its encoder has attention on full sequence \cite{chen2020developing}. However, it requires access to the full sequence, and the computational cost grows quadratically with respect to the input sequence length. These factors limit its adoption for streaming applications \cite{wu2020streaming}. 
Additionally, transformers' high memory consumption and inference time pose practical difficulties for deploying and updating large-scale models. 

Self-attention mechanism in transformers has quadratic complexity with respect to sequence length, limiting scalability for long sequences. To address this issue, several solutions, such as sparse attention patterns \cite{zhao2022adaptive,woo2021speech,avinava2021constructing}, low-rank factorization \cite{winata2020lightweight}, random feature maps \cite{krzysztof2020rethinking}, and locality-sensitive hashing \cite{kitaev2020reformer,woo2021speech,roy2021efficient}, have been proposed \cite{tay2022efficient}. The memory consumption of transformers grows linearly with sequence length and quadratically with hidden dimension size, creating challenges for large-scale data training and inference. Some solutions to this problem include reversible residual connections \cite{yu2022auxiliary,duan2022dual}, gradient checkpointing \cite{walmart2021sparse}, weight sharing \cite{xiao2019sharing,han2021connection}, or parameter pruning \cite{li2021differentiable,li2019improving} to save memory. Chen et al. \cite{chen2021developing} also show the use of streaming processing and early stopping to reduce latency and run-time cost in speech models. Lin et al. \cite{lin2022compressing} used weight pruning, head pruning, low-rank approximation, and knowledge distillation to reduce parameters. Parallelizing and accelerating transformer models on different hardware platforms may encounter challenges such as load imbalance, communication overhead, or memory fragmentation \cite{shi2021emformer}. Several solutions have been proposed to improve hardware utilization, including tensor decomposition \cite{ma2019tensorized,gu2022heat,pham2022tt,li2022hypoformer}, kernel fusion \cite{denis2022accelerated}, mixed precision arithmetic \cite{xu2021mixed}, or hardware-aware optimization \cite{wang2020hat,kuchaiev2018mixed}.


Efficiency is another major concern for Transformers due to their large and complex architectures. When these models are pre-trained, they may not be efficient for all downstream tasks due to different data distributions. To improve efficiency, recent efforts have attempted to find solutions to use fewer training data and/or parameters. These solutions include knowledge distillation, simplifying and compressing the model, using asymmetrical network structures, improving utilization of training samples, compressing and pruning the model, optimizing the complexity of self-attention, optimizing the complexity of self-attention-based multimodal interaction/fusion, and optimizing other strategies. Several specific methods have been proposed to address these issues, including knowledge distillation by Miech et al. \cite{miech2021thinking} and Touvron et al. \cite{touvron2021training}, model simplification by Xu et al. \cite{xu2021e2e}, Kim et al. \cite{kim2021vilt}, and Akbari et al. \cite{akbari2021vatt}, weight-sharing by Wen et al. \cite{wen2021cookie} and Lee et al. \cite{lee2020parameter}, training with fewer samples by Li et al. \cite{li2021supervision}, compressing and pruning the model by Gan et al. \cite{gan2022playing}, optimizing the complexity of self-attention by Child et al. \cite{child2019generating} and Transformer-LS \cite{gan2021transformerls}, optimizing the complexity of self-attention based multimodal interaction/fusion by Nagrani et al. \cite{nagrani2021attention} and optimizing other strategies by Yan et al. \cite{yan2022multiview}. These efforts demonstrate the importance of addressing efficiency in the development of Transformers.

\subsection{Large Data Requirements}


One significant challenge faced by transformers-based speech models is the requirement for a large amount of data for effective training. 
While recent text-based conversational AI models have been trained on large amounts of data (e.g., GODEL, which used around 800GB of data, and GPT-3, which was pretrained on 570GB of data), the amount of speech data available for training models for spoken technology is more limited. For instance, the works in Table~\ref{table:sds-transformers} (with some exceptions) use datasets consisting of several thousand spoken dialogues for various tasks, which is far less compared to the hundreds of gigabytes of text data used by text-based models. This highlights the need to either develop more efficient ways of creating datasets to train speech-related systems more efficiently.

One approach to enhancing the performance of transformer-based models for speech recognition is to collect a large-scale dataset of multilingual and multitask audio data from various sources \cite{radford2022robust}. This dataset can then be used to train a transformer-based model with self-attention layers using weak supervision. Evaluating the model on various downstream tasks and benchmarks can further improve its performance. Additionally, data augmentation and transfer learning can also be used to improve model performance \cite{chen2022wavlm}. Data augmentation techniques such as pitch shifting, time stretching, noise injection, SpecAugment, etc.,  can be employed to increase the diversity and robustness of the training data. Another solution is to use pre-trained models by training them on learning to learn generalised representation from large-scale unlabeled data \cite{ma2020data} \cite{park2019specaugment}. These models can be fine-tuned using few-shot learning to get better performance on downstream tasks. 
Multi-task learning approaches can also be utilized  to enhance the performance of transformer-based models for speech and language processing with smaller datasets \cite{chen2022wavlm}. This includes masked acoustic modeling, contrastive predictive coding, speaker classification, emotion recognition, and sentiment analysis. 


\subsection{Generalization and Transferability}
Transformers face challenges with generalization and transferability that can affect their ability to handle a broader range of tasks and scenarios. One of the main issues with generalization is the absence of inductive biases in pure transformers, which makes them heavily reliant on large-scale training data for optimal performance \cite{dosovitskiy2020image}. This can lead to poor performance on downstream tasks if the training data quality is poor. Additionally, transformers lack built-in biases, unlike convolutional neural networks, which makes it more difficult for them to generalize to new tasks or scenarios \cite{dosovitskiy2020image}. 
To address the challenge of poor generalization on new domains, Xue et al. \cite{xue2021bayesian} proposed the Bayesian Transformer Language Model (BTLM), which integrates a Bayesian framework to enhance the model's ability to handle out-of-domain data.  Bayesian Transformer \cite{xue2021bayesian} uses variational inference to estimate the latent parameter posterior distributions and account for model uncertainty while another model \cite{lin2022compressing} resolves generalization issue by applying several compression techniques, such as weight pruning, head pruning, low-rank approximation, and knowledge distillation, to a 12-layer Transformer model trained with contrastive predictive coding (CPC). This approach delivers improved performance on out-of-domain data compared to traditional language models. Other proposed solutions include Parallel Scheduled Sampling (PSS) and Relative Positional Embedding (RPE) \cite{zhou2019improving}. PSS improves robustness and reduces exposure bias by randomly sampling tokens from either ground truth or predicted sequences during training. RPE encodes relative distances between tokens rather than absolute positions, which enhances the modeling capability for long sequences. Various other papers also discuss the issues and solutions of generalisation in transformer-based speech models along with various solutions. Zhou et al. \cite{zhou2019improving} propose a text-to-speech model, GenerSpeech that transfer unseen style features from an acoustic reference to a target text. It improves generalization by decomposing the speech variation into style-agnostic and style-specific parts, and by using a content adaptor with Mix-Style Layer Normalization to eliminate style information in the linguistic content representation.

Transferability is also a significant challenge for transformers due to domain gaps.  Several methods have been proposed to enhance the transferability of  transformers, including data augmentation, adversarial perturbation strategies, learning a shared embedding space, and knowledge distillation \cite{gan2020large,radford2021learning,kervadec2021supervising}. While these methods have shown promising results, there are still some obstacles to transferability for multimodal applications. One of the challenges is the distribution gap between training data and practical data, as shown by Zhan et al. \cite{zhan2021product1m}. They demonstrate that transferring supervised multimodal transformers pre-trained on well-aligned cross-modal pairs/tuples to weakly aligned test data is challenging. Rahman et al. \cite{rahman2020integrating} and Xia et al. \cite{xia2021xgpt} show that transferring multimodal transformers across different tasks requires careful adaptation and fine-tuning. In multi-language data settings, transformers also face transferability challenges as demonstrated by Zhou et al. \cite{zhou2021uc2} and Ni et al. \cite{ni2021m3p}. 



\subsection{Multimodal Training}
In Multimodal Learning (MML) Transformers, a fusion of information across multiple modalities is typically achieved at three conventional levels: input (i.e., early fusion), intermediate representation (i.e., middle fusion), and prediction (i.e., late fusion) \cite{chen1998audio}. Middle fusion can be achieved by directly feeding the representations of two modalities into the standard attention module, which is followed by latent adaptation and ends up with a late fusion of final bimodality representations \cite{tsai2019multimodal,sahu2020low}. This idea can be extended by alternating or compounding with unimodal attention, or token exchange across modalities \cite{gao2019dynamic,li2021causal,chen2021crossvit}. On the other hand, inspired by the success of BERT, different modalities can integrate as early as at the input stage \cite{sun2019videobert,li2019visualbert,chen2020uniter,li2020unicoder,li2020oscar,huang2020pixel,zhu2020actbert,qi2020imagebert,nagrani2021attention,zhuge2021kaleidobert}. These models are known as one-stream architecture, which allows the adoption of the merits of BERT with minimal architectural modification. However, a major difference with these one-stream models is the usage of problem-specific modalities with variant masking techniques. A noticeable fusion scheme is introduced based on a notion of bottleneck tokens, which applies for both early and middle fusion by simply choosing to-be-fused layers \cite{nagrani2021attention}. Late fusion based on prediction is less adopted in MML Transformers as the focus is on learning stronger multimodal contextual representations \cite{chen1998audio,owens2018audio}. The interaction between modalities can be explored further for enhancing and interpreting the fusion of MML \cite{liu2021probing}.

Another issue with multimodal transformer models in real-world scenarios is that the data often exist in multiple modalities with intrinsic synchronization, such as audio-visual correspondence \cite{morgado2020learning}, which forms the basis for cross-modal alignment. Recently, there has been a surge of interest in leveraging large quantities of web data (e.g., image-text pairs) for vision and language tasks using Transformers-based alignment \cite{radford2021learning,jia2021scaling,xu2021videoclip,lei2021less}. The approach involves mapping two modalities into a common representation space with contrastive learning over paired samples and is typically implemented using massive multimodal models (MML) that are expensive to train. As a result, subsequent works have focused on utilizing pre-trained models for various downstream tasks \cite{wang2022clip,li2022align,luo2022clip4clip,fang2021clip2video,narasimhan2021clip}. These alignment models are capable of zero-shot transfer, particularly for image classification via prompt engineering, which is remarkable given that image classification is traditionally considered a uni-modal learning problem, and zero-shot classification remains a difficult challenge despite extensive research \cite{xian2018zero}. The approach has also been extended to more challenging and fine-grained tasks, such as object detection \cite{gu2021open}, visual question answering \cite{tan2019lxmert,chen2020uniter,li2020oscar,liu2021align}, and instance retrieval \cite{liu2021gilbert,liu2021align}, by introducing region-level alignment, which incurs additional computational costs from explicit region detection. TEASEL \cite{arjmand2021teasel} uses a speech-prefixed language model that takes speech features as input and predicts masked tokens in text and resolves issues with multimodal training.

\subsection{Robustness}

Despite their widespread adoption in speech processing applications, transformers exhibit sensitivity to domain shifts and noise in speech data leading to a sub-optimal performance in downstream tasks. Additionally, these models may not generalize well to other languages when trained solely on monolingual data~\cite{kawakami2020learning}.
This issue has been identified as one of the causes of performance degradation in transformer-based ASR system~\cite{burchi2023audio}, speech-to-animation models~\cite{novoselov2022robust}, speaker recognition and speech-to-speech translation models~\cite{jia2021translatotron} as raw speech features used as input render these models sensitive to noise and speaker variations. Moreover, the performance of these models is negatively impacted by the lack of prosody information, as they do not explicitly model it.

Several solutions have been proposed to overcome these challenges. One approach is to learn robust and multilingual speech representations using contrastive learning, which is a self-supervised technique that encourages the model to distinguish between similar and dissimilar speech segments. A multilingual phonetic vocabulary is used to capture cross-lingual similarities and enable transfer learning across languages. Additionally, self-training and semi-supervised learning are applied to leverage unlabeled data and enhance the quality of the representations. These approaches have been shown to outperform previous methods on various benchmarks and achieve state-of-the-art results on low-resource speech recognition \cite{kawakami2020learning}. To address the challenges associated with noisy or distorted speech signals, a novel audio-visual ASR model has been proposed, leveraging both speech and lip movement information to improve recognition accuracy and robustness. The model is based on the Efficient Conformer architecture, which combines convolutional neural networks. To overcome the challenges faced by speaker recognition models, an unsupervised approach that uses a contrastive loss to learn speaker embeddings directly from raw speech signals has also been proposed. The model uses a multi-task learning approach, where both speaker recognition and ASR tasks are learned jointly. This approach can help to overcome the variability in speech signals caused by different speakers and speaking styles, thereby improving the robustness of the ASR system.


\section{Summary and Conclusions} \label{sec:conclusion}

The transformer architecture has emerged as a highly effective neural network architecture in the field of speech processing due to its ability to handle sequential data for various speech-related tasks. The popularity of transformers has been further accelerated by the availability of specialized libraries for transformer-based speech-processing tasks. The key innovation of transformers lies in their ability to capture long-range dependencies among input sequences using self-attention layers, and its effectiveness has been demonstrated in various speech processing tasks such as automatic speech recognition, text-to-speech synthesis, speaker recognition, multi-microphone processing, and speech translation. 

This pioneering paper presents the first detailed and comprehensive survey of the applications of transformers in the audio domain. Our review shows that transformers are increasingly becoming popular in the speech-processing community. The main findings of this paper are summarized below.

\begin{itemize}
\item Transformers provide a competitive alternative to Recurrent Neural Network (RNN)-based models in Speech Processing tasks and have shown promising results in Automatic Speech Recognition (ASR) and Text-to-Speech (TTS). Transformers use self-attention layers to capture long-range dependencies among input sequences, allowing more parallelization than RNNs.
\item Pre-training with self-supervised learning techniques like wav2vec and data2vec can lead to faster convergence and performance boost in transformers.
\item Hybrid models that combine transformers with conventional acoustic modeling techniques can improve word error rates and reduce computational complexity.
\item Attention should be paid to overfitting and generalization problems in transformer models in ASR and TTS.
\item Different modeling units such as syllables and phonemes should be compared in ASR using transformers.
\item Multi-head attention mechanisms in transformers can improve parallelization by solving the long-distance dependency problem in end-to-end TTS architectures.
\item Length regulator based on duration predictor can be used to solve the issue of sequence length mismatch in TTS.
\item Data augmentation and contrastive learning techniques can be used to build cross-lingual or multilingual speech recognition systems using pre-trained transformer models.
\item Ethical concerns around privacy, bias, and fairness should be considered when developing and deploying speech processing systems using transformers.
\item Robustness of transformer models to adversarial attacks and out-of-distribution inputs should be carefully evaluated and addressed in ASR and TTS applications.

\end{itemize}

Future research on cross-lingual/multilingual systems is needed to address the performance issues highlighted in this review. These recommendations are intended for researchers and developers in the field of speech processing. 



\section{Acknowledgements}
We would like to thank Fatima Seemab (NUST) for initially working on this paper.



\end{document}